\pdfoutput=1

\documentclass[11pt]{article}

\usepackage[final]{acl}
\usepackage{algorithm}
\usepackage{algorithmic}
\usepackage{amssymb}
\usepackage{amsmath}
\usepackage{multirow}
\usepackage{times}
\usepackage{latexsym}

\usepackage[T1]{fontenc}

\usepackage[utf8]{inputenc}

\usepackage{microtype}

\usepackage{inconsolata}

\usepackage{graphicx}

\title{My Words Imply Your Opinion: Reader Agent-based Propagation Enhancement for Personalized Implicit Emotion Analysis}

\author{
	\textbf{Jian Liao\textsuperscript{\rm 1,4}},
	\textbf{Yu Feng\textsuperscript{\rm 1}},
	\textbf{Yujin Zheng\textsuperscript{\rm 1}},
	\textbf{Jun Zhao\textsuperscript{\rm 2}},
	\textbf{Suge Wang\textsuperscript{\rm 1,3}$^*$},
	\textbf{Jianxing Zheng\textsuperscript{\rm 1}}
	\\
	\textsuperscript{\rm 1}School of Computer and Information Technology, Shanxi University, China\\
	\textsuperscript{\rm 2}Institute of Automation, Chinese Academy of Science, China\\
	\textsuperscript{\rm 3}Key Laboratory of Computational Intelligence and Chinese Information Processing \\of Ministry of Education, Shanxi University, China\\
	\textsuperscript{\rm 4}Joint Laboratory of Tourism Big Data in Shanxi Province, China
	\\	
	\small{
		\textbf{$^*$Correspondence:} \href{mailto:wsg@sxu.edu.cn}{wsg@sxu.edu.cn}
	}
}

\begin{document}
\maketitle
\begin{abstract}
The subtlety of emotional expressions makes implicit emotion analysis (IEA) particularly sensitive to user-specific characteristics. Current studies personalize emotion analysis by focusing on the author but neglect the impact of the intended reader on implicit emotional feedback. In this paper, we introduce Personalized IEA (PIEA) and present the RAPPIE model, which addresses subjective variability by incorporating reader feedback. In particular, (1) we create reader agents based on large language models to simulate reader feedback, overcoming the issue of ``spiral of silence effect'' and data incompleteness of real reader reaction. (2) We develop a role-aware multi-view graph learning to model the emotion interactive propagation process in scenarios with sparse reader information. (3) We construct two new PIEA datasets covering English and Chinese social media with detailed user metadata, addressing the text-centric limitation of existing datasets. Extensive experiments show that RAPPIE significantly outperforms state-of-the-art baselines, demonstrating the value of incorporating reader feedback in PIEA. The code and datasets of this study are available at \url{https://github.com/sxu-nlp/RAPPIE}. 
\end{abstract}

\section{Introduction}
\begin{figure}[t]
	\centering
	\includegraphics[width=0.98\linewidth]{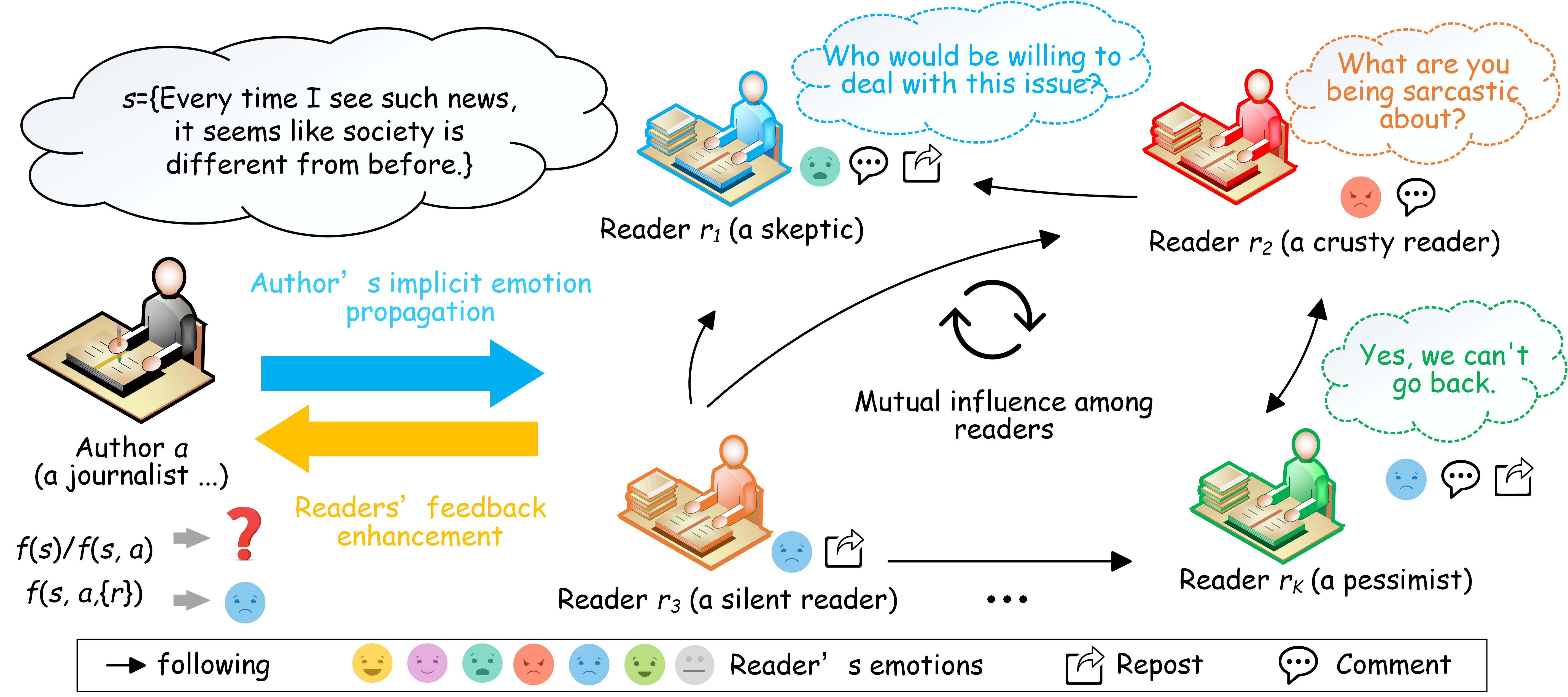} 
	\caption{The illustration of the implicit emotion dynamic propagation, interaction, and enhancement between author and readers in social media.}
	\label{fig_challenges}
\end{figure}

In recent years, the field of natural language processing has seen a surge in interest towards Implicit Emotion Analysis (IEA), which is focused on discerning emotions from objective statements that lack explicit emotional expressions but still convey such feelings \cite{LIAO2022}.
Unlike classical emotion analysis, which relies on explicit sentimental cues and can be formalized as $f(s)$ (where a model $f$ predicts the emotion of a given content $s$), IEA is highly subjective and influenced by user traits. This leads to diverse emotional responses to the same content, necessitating the refinement to Personalized IEA (PIEA) as $f(s,a)$, where $a$ represents the author of $s$.
For the example \textit{s} in Figure~\ref{fig_challenges}, the emotion expressed by a pessimist or a skeptic can be quite distinct.

Current studies on sentiment/emotion analysis that consider user subjectivity usually introduce author's information into the text by incorporating the author's attribute \cite{mireshghallah2022,liao2023} or author-product interaction relationships \cite{kertkeidkachorn2023,lyu2023}.
However, these efforts often overlook the potential influence of content readers on implicit emotions. This is particularly evident in the social media scenario, where some authors may deliberately employ ambiguous emotional expressions with provoking or confrontational overtones when composing their content, aiming to elicit predetermined emotions from their readership, see Figure~\ref{fig_challenges} for an example. According to the theory of the implied reader in narratology, the implicit emotions in the author's content are not generated by the readers but rather conveyed to the readers through text, narrative, and plot during the creative process of the author \cite{Schmid2010}. Therefore, in this study, we propose to model the reader's feedback and interactive propagation process and integrate them into the author's personalized implicit emotion modeling, formalized as $f(s,a,\{r\})$, where $\{r\}$ is the reader group of $s$.

To the best of our knowledge, there has been no research that utilizes readers' information to enhance authors' implicit emotion identification. This is primarily due to the challenges faced in the modeling of reader's feedback and interactive propagation process, summarized in the following:

(1) \textbf{How can we obtain readers' feedback towards the author's content?} Although the commentary behaviors of readers on social media theoretically can be captured through the crawling of forwarding chains, the ``spiral of silence effect'' \cite{Sohn2022Spiral} dictates that most readers do not engage in observable actions (like reposting or commenting). Besides, the forwarding chains are subject to dynamic changes due to reader actions like deletion or revision, which hampers the accurate and comprehensive acquisition of real reader feedback data. Fortunately, existing agent-based research \cite{pmlr-v202-aher23a,xie2024can} demonstrates that Large Language Models (LLMs) can effectively simulate human behavior, supporting the construction of user agents to model their emotional feedback in social media scenario.

(2) \textbf{How can the interactive propagation among readers be modeled?} Different readers may adopt varying emotions and behaviors upon engaging with the content from an author, and within the reader community, they assume different propagation roles in the process of interaction. This commonality of propagation roles provides a certain supplementation when reader information is inadequate \cite{li2024your}. 
To capture the intricate dynamics among readers, a role-aware Graph Neural Network (GNN) serves as an optimal method \cite{zhao2024rdgcn}. Incorporating role information into graph propagation learning enables precise analysis of the emotional states induced by the author's implicit emotions within the reader community.
This facilitates a comprehensive understanding of how the interactive propagation among readers with distinct roles influences overall feedback.

(3) \textbf{There is a lack of high-quality implicit emotion datasets with user information available.}
Existing IEA datasets only focus on annotating the emotion of textual content, but lack user-related information such as historical postings and social relationships. This limitation results in existing methods resorting to fictional user identifiers for personalized analysis \cite{mireshghallah2022}. \citet{liao2023} constructed a Chinese dataset for personalized sentiment analysis, but the annotations are limited to coarse-grained sentiment polarities.

In this paper, we introduce the PIEA task which incorporates user-specific information to address the subjective variability in classical IEA. To overcome the scarcity of user-specific data in existing corpora, we construct two PIEA datasets covering both Chinese and English, encompassing implicit emotions, user attributes, social relationships, and historical postings.
We propose a novel model named \textbf{R}eader \textbf{A}gent-based \textbf{P}ropagation enhancement for \textbf{P}ersonalized \textbf{I}mplicit \textbf{E}motion analysis (RAPPIE). RAPPIE first utilizes LLM to create reader agents that simulate readers' emotional feedback to authors' content. Drawing on communication theory, we categorize reader roles into four types and develop a role-aware multi-view interactive propagation graph model to capture the interactive propagation among readers from different behavioral perspectives.
Then we integrate author attributes and fused readers' feedback with the content to enhance PIEA. 

Our key contributions include:

(1) Refining IEA into PIEA and proposing the RAPPIE model, which effectively models authors' implicit emotions through agents-based simulation of reader's emotions and behaviors, mitigating issues related to the ``spiral of silence effect'' and behavior data incompleteness in acquiring real reader feedback.

(2) Developing a role-aware multi-view interactive propagation graph learning method to address reader information sparsity and model the emotion propagation and interaction among readers.

(3) Introducing two new PIEA datasets covering different linguistic and cultural backgrounds, enriched with comprehensive user metadata beyond textual annotations. Extensive experiments validate the superior performance of our approach compared to state-of-the-art (SOTA) baselines.

\section{Related Work}
\subsection{Implicit Sentiment Analysis}
The interdependence between emotion and sentiment has been well-recognized. Recent research has mostly focused on implicit sentiment analysis, mainly categorized into deep semantic modeling-based and knowledge enhancement-based approaches.

Deep semantic modeling methods typically leverage textual structural information to capture implicit sentiment expressions, such as aspect-category-opinion-sentiment quadruples \cite{cai2021}, alignment of implicit sentiment-emotion label \cite{li2021}, causal intervention between sentence and sentiment \cite{wang2022causal}, and dependency measurement in both distance and type views \cite{zhao2024rdgcn}. \textbf{These methods still struggle with insufficient emotional information within the text itself.}
On the other hand, knowledge enhancement approaches, such as incorporating explicit commonsense knowledge \cite{LIAO2022}, user attribute information \cite{lyu2023,liao2023}, abstract meaning representation graphs \cite{ma2023,tran2023}, and chain-of-thought reasoning based on LLM \cite{fei2023}, aim to interpret implicit emotions by integrating external knowledge. Nevertheless, \textbf{the external knowledge sources introduced by these methods are still limited to the content authors themselves}, making it challenging to enhance the implicit emotion identification process with readers' emotional feedback.

\subsection{LLM-based Agent}
In recent years, the expansion of model scale and the abundance of training data have empowered LLMs with unprecedented capabilities. Research by \citet{guo2024integrating} and \citet{wang2024deem} has demonstrated the ability of large language models to simulate human behaviors from various perspectives, particularly showcasing unique strengths in language comprehension and generation. Researchers in diverse fields have also taken advantage of the powerful semantic understanding ability of LLMs to design intelligent agents that address a variety of complex application scenarios, such as fake news detection \cite{nan2024let}, recommendation systems \cite{huang2024large}, and dialogue domains \cite{owoicho2023exploiting,davidson2023user,li2024your}. Nevertheless, these approaches construct LLM-based agents primarily through behavioral simulation, \textbf{overlooking the interactions and representation modeling among multiple agents influenced by their respective propagation roles}.

\section{Method}

\subsection{Problem Definition and Notations}
\label{define}
Given a predefined pair $\langle s, u \rangle$, where $s$ represents implicit emotional text, authored by a user \textit{u}. PIEA aims to identify the potential implicit emotions and their categories in \textit{s}, based on \textit{u}'s personalized information. A user \textit{u} is described by: user attributes $u_a$ and historical posts $u_h$. $u_a$ primarily encompass individual information such as gender $u_a^g$, region $u_a^r$, and personalized tag $u_a^t$, while $u_h$ collects other content posted by \textit{u}. Additionally, a user relational network graph $G_f$ is also included to provide the \textit{following} relationships between users.

\subsection{RAPPIE Model}
\begin{figure*}[ht]
	\centering
	\includegraphics[width=0.99\textwidth]{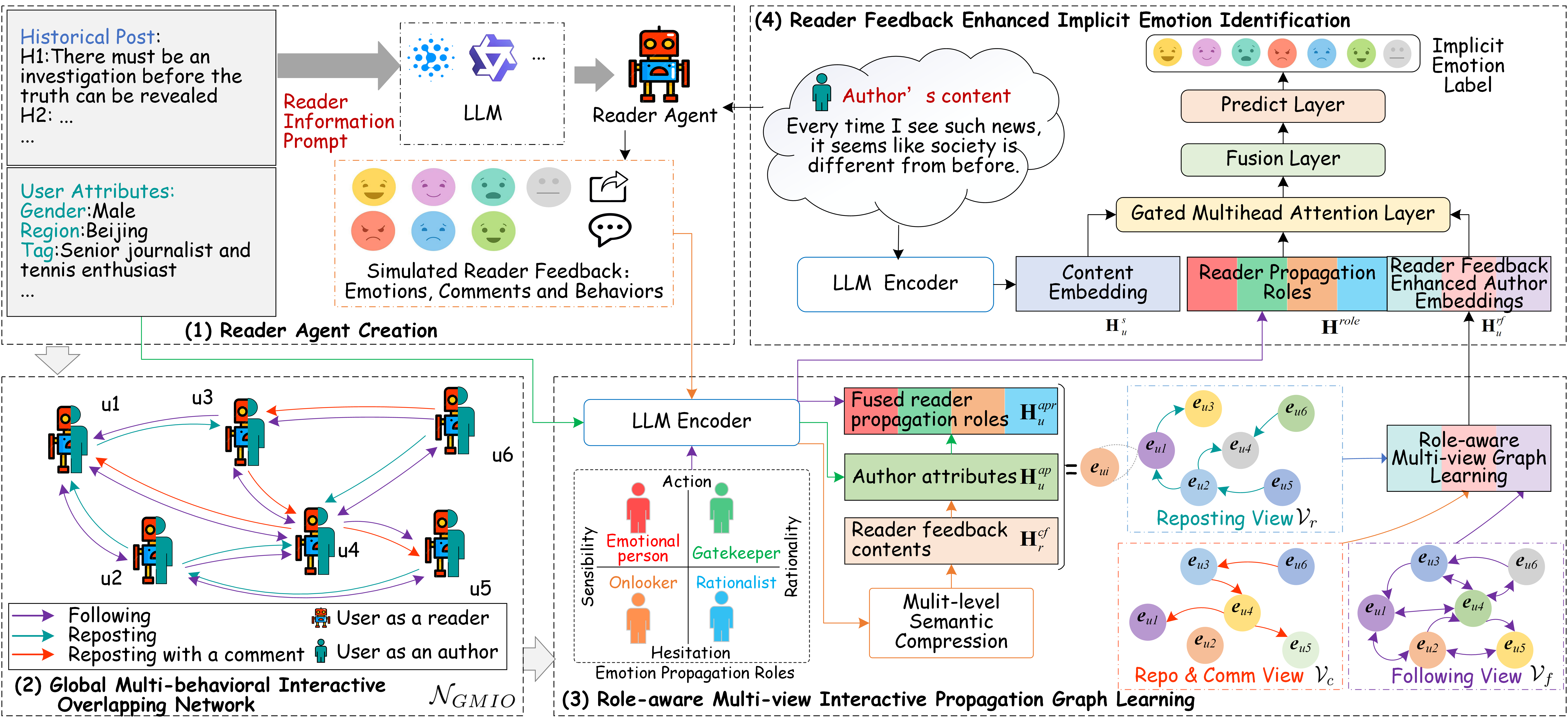} 
	\caption{The framework of our proposed RAPPIE model. The entire study consists of four core modules: (1) utilizing reader information prompts to construct the reader agent for feedback simulation, (2) a global interactive multi-behavior overlapping network is constructed based on the simulated behaviors of \textit{reposting} and \textit{reposting with a comment}, as well as \textit{following}, among different users, (3) a role-aware multi-view GNN is employed to learn users representations from multiple interactive perspectives, integrating reader propagation role during the graph learning process, (4) implicit emotion prediction is carried out by fusing content semantic with propagation role features and reader-feedback enhanced author embeddings.}
	\label{fig_framework}
\end{figure*}

The framework of our proposed RAPPIE model is illustrated in Figure~\ref{fig_framework}. 

\subsubsection{Reader Agent Creation}
Given a set of users $\{u\}$, we construct prompt templates for creating reader agents based on user attributes $u_a$ and user historical posts $u_h$. We randomly sample content from users' posts in the training set to build the prompts, as illustrated in Figure~\ref{fig_prompt_agent} in Appendix~\ref*{agentprompt}. Considering the agent creation process, we employ multiple LLMs with different architectures as the foundations for reader agents, which are used to generate readers' feedback on a given post, including emotions, comments, and behaviors.

\subsubsection{Global Multi-behavioral Interactive Overlapping Network}
\label{sec:global}
Based on the created reader agent, we construct an overlapping network, denoted as $\mathcal{N}_{GMIO}=\{U, \{E_{fw}, E_{rp}, E_{rc}\}\}$, that captures the global multi-behavioral interactions among users. Each node $u\in U$ represents a user, who assumes two identities: as an author of their content and as a reader of others' posts. Multiple heterogeneous directed edges may exist between a pair of nodes, including three types of interactive behaviors: \textit{following}, \textit{reposting}, and \textit{reposting with a comment}.

The following edges $E_{fw}$ can be directly obtained from the graph $G_f$, and we utilize reader agents to simulate the other two behaviors. For each user node $u\in U$, we first acquire the set of posts $P_u$ that user \textit{u} have published as authors within the training set. We then construct prompts based on $\langle p, r \rangle$ pairs ($p\in P_u$, $r\in U$ and $r \neq u$) following Figure~\ref{fig_prompt_agent} in Appendix~\ref{agentprompt}, and input them into the reader \textit{r}'s agent. It's worth noting that this simulation process is not limited to readers who follow \textit{u}, which helps in obtaining more comprehensive reader feedback but leads to a huge number of pending $\langle p, r \rangle$ pairs.
To address this, we employ a text similarity calculation method based on \textit{Tf-Idf} to filter the top-\textit{k} posts ${P_r}^{\text{top}-k}$ semantically relevant to the reader \textit{r}'s historical posts $u_h^r$, and use ${P_r}^{\text{top}-k}$ for feedback simulation to establish edges $E_{rp}$ and $E_{rc}$ between nodes \textit{r} and \textit{u}, respectively, see Appendix~\ref{readerback} for detailed simulation results.

\subsubsection{Role-aware Multi-view Interactive Propagation Graph Learning}

\textbf{\quad (1) Multi-view generation} 

Given the diverse heterogenous relationships represented by the edges between nodes in the overlapping network $\mathcal{N}_{GMIO}$ previously constructed, adopting a multi-view learning approach to capture the mutual influences under each interactive relationship or behavior and then fusing these representations is a promising solution \cite{multiview}. 
In this section, we propose a role-aware multi-view interaction propagation graph learning method. Based on the overlapping network $\mathcal{N}_{GMIO}$ constructed in the previous section, we generate three independent views of user interactive relationships focusing on \textit{following}, \textit{reposting}, and \textit{reposting with a comment}, denoted as $\mathcal{V}_f$, $\mathcal{V}_r$, and $\mathcal{V}_c$. For a user \textit{u}'s node in the graph, the node embedding $\mathbf{e}_u$ consists of three parts: the attributes associated with \textit{u} when acting as an author, the simulated feedback content when \textit{u} is a reader, and the reader propagation role information.

\textbf{(2) Agent-generated comments encoding}

For simulated reader feedback modeling, given a related post $P_r^i \in {P_r}^{\text{top}-k}$ to reader agent $r$, the generated emotions and comment content (with a <CLS> token added at the beginning) in $r$'s feedback is denoted as $c_r^i$. An LLM-based encoder is employed to obtain the initial semantic encoding. In Particular, we extract the input matrix of the final output layer, which predicts the probability distribution of the generated sequence as the semantic encoding matrix of the input prompt, defined as $\mathbf{Mc}_r^i = LLM(c_r^i) \in \mathbb{R}^{N\times d_1}$, $N$ is the unified maximum length of generated content.
Then, a 2-layer self-attention mechanism is applied for semantic compression on $\{\mathbf{Mc}_r^i\}$, $i \in k$, defined as:
\begin{equation}\label{Eq_sa1}
	\tilde{\mathbf{Mc}}_r = \{\text{Self-att}(\mathbf{Mc}_r^i)\}_{i \in k}
\end{equation}
\begin{equation}\label{Eq_sa2}
	\mathbf{H}_r^{cr} = \text{Self-att}(\tilde{\mathbf{Mc}}_r)
\end{equation}
Eq.\eqref{Eq_sa1} integrates the semantics within $\mathbf{Mc}_r^i$ and extracts the embedding of the <CLS> token as the semantic encoding for $c_r^i$. Eq.\eqref{Eq_sa2} further consolidates all $c_r^i$ and uses the first <CLS> encoding from the fused embeddings as the compressed semantic feature $\mathbf{H}_r^{cr} \in \mathbb{R}^{d_1}$ of agent-generated comments.

\textbf{(3) Author attributes encoding}

For author attributes, we design a prompt template $prompt_u$ as ``\textit{User individual information}=$[$\textit{gender}:$u_a^g$, \textit{region}:$u_a^r$, \textit{tag}:$u_a^t]$'' and input the filled prompt into the LLM-based encoder with a pooling layer, denoted as Eq.\eqref{Eq_llm}. 
\begin{equation}\label{Eq_llm}
	\mathbf{H}_u^{at} = \text{Pooling}(LLM(prompt_u)) \in \mathbb{R}^{d_1}
\end{equation} 
where $d_1$ is the dimension of embedding, and Pooling($\cdot$) is the average pooling operation for dimensionality compression of the LLM-based encoded matrix.
Then $\mathbf{H}_u^{at}$ is fused with the agent-generated comment embedding $\mathbf{H}_r^{cr}$ obtained by Eq.\eqref{Eq_sa2} when user \textit{u} is identified as a reader, defined as Eq.\eqref{Eq_hua}.
\begin{equation}\label{Eq_hua}
	\mathbf{H}_u^{ap} = \mathbf{W}_1(\mathbf{H}_u^{at}\oplus \mathbf{H}_r^{cr})+\mathbf{b}_1
\end{equation} 
where $\mathbf{W}_1\in \mathbb{R}^{d_1\times 2d_1}$, and $\mathbf{b}_1 \in \mathbb{R}^{d_1}$ are the trainable parameters, and $\oplus$ means the operation of concatenation.

\textbf{(4) Propagation roles encoding}

In communication studies, propagation roles reflect the shared behavioral preferences of a user category, offering essential supplementation when user data is sparse. In this study, drawing upon the role behavior theory \cite{biddle2013role} in communication, we establish a quaternary role system based on the dimensions of users' rationality-sensibility and action-hesitation, and define four propagation roles: \textit{Emotional person}, \textit{Gatekeeper}, \textit{Onlooker}, and \textit{Rationalist}.
Additionally, we craft role descriptions, grounded in specific events, which provide foundational information for role modeling, a more detailed explanation of the propagation roles can be found in Appendix~\ref{proprole}.
To capture the propagation roles' features, we concatenate each role's description with example content and feed them into an online LLM. By leveraging the LLM's background knowledge, we obtain a more enriched role explanation, denoted as $r_{extend}^i$. A rewriting prompt ``\textit{Please repeat the following content}: [$r_{extend}^i$]'' is designed to acquire the semantic encodings $\mathbf{R}_i$ ($i\in \{E,G,O,R\}$) of the propagation roles through the LLM-based encoder. The feature embeddings of the four propagation roles constitute a role semantic matrix denoted as $\mathbf{H}^{role}\in \mathbb{R}^{4\times d_1}$. To model the associative distribution between users and roles, we employ a multi-head attention mechanism (MATT for short) to integrate the semantic features of propagation roles with the fused user embedding $\mathbf{H}_u^{ap}$, defined as Eqs.\eqref{Eq_Role} and \eqref{Eq_userfusion}.
\begin{equation}\label{Eq_Role}
	\mathbf{M}_u^{apr} = \text{MATT}(\mathbf{H}_u^{ap}, \mathbf{H}^{role}, \mathbf{H}^{role})
\end{equation} 
\begin{equation}\label{Eq_userfusion}
	\mathbf{H}_u^{apr} = \mathbf{W}_2(\mathbf{H}_u^{ap}\oplus \mathbf{M}_u^{apr})+\mathbf{b}_2
\end{equation} 
where $\mathbf{W}_{2}\in\mathbb{R}^{d_1\times 2d_1}$, $\mathbf{b}_{2}\in\mathbb{R}^{d_1}$ are the trainable parameters.

\textbf{(5) Multi-view graph learning}

For each view $v\in \{\mathcal{V}_f,\mathcal{V}_r,\mathcal{V}_c\}$, the node embedding $\mathbf{e}_u$ of user \textit{u} is initialized with $\mathbf{H}_u^{apr}$, and the LightGCN \cite{he2020lightgcn} is adopted to model the interaction influence between users. The graph propagation learning and optimization processes are defined as Eqs.\eqref{Eq_lightgcn}-\eqref{Eq_gcnopt}.
{\setlength\abovedisplayskip{3pt}	
	\setlength\belowdisplayskip{3pt}
\begin{equation}\label{Eq_lightgcn}
	\mathbf{e}_u^{v(T+1)}=\frac{1}{|\mu_u^v|}\sum_{j\in \mu_u^v} \mathbf{e}_j^{v(T)}
\end{equation} 
\begin{equation}\label{Eq_gcnopt}
	\mathbf{\tilde{y}}_{ij}=\text{softmax}(\mathbf{W}_v(\mathbf{e}_i^{v(T)}\oplus \mathbf{e}_j^{v(T)}) +\mathbf{b}_v)
\end{equation}}where $|\mu_u^v|$ means the size of the first-order neighbor set $\mu_u^v$ for user $u$ in the given view $v$, $T=\{0,1,\dots,K\}$ denotes the number of propagation rounds. $i$, $j\in U$ and $i\ne j$. $\mathbf{W}_v\in \mathbb{R}^{2\times2d_1}$ and $\mathbf{b}_v\in \mathbb{R}^2$ are the trainable parameters. 

The cross-entropy loss function is employed to optimize the model by minimizing the difference between the predicted edge distribution $\mathbf{\tilde{y}}_{ij}$ and the golden label $\mathbf{y}_{ij}$. After all views have been optimized, the node embedding $\mathbf{e}_u^{v}$ of user \textit{u} across these views collectively constitute the matrix $\mathbf{H}_u^{rf}=[\mathbf{e}_u^{\mathcal{V}_f}, \mathbf{e}_u^{\mathcal{V}_r}, \mathbf{e}_u^{\mathcal{V}_c}]\in \mathbb{R}^{3\times d_1}$, which embedded reader feedback information.

\subsubsection{Reader Feedback Enhanced Implicit Emotion Identification}
To perform information enhancement on implicit emotional contents, we also obtain the semantical matrix of implicit emotional content \textit{s} published by \textit{u} using the LLM-based encoder, defined as $\mathbf{H}_u^{s} = LLM(s) \in \mathbb{R}^{N\times d_1}$, where \textit{N} is the unified maximum length of input texts.
The gated multi-head attention model is employed to fuse the reader-feedback enhanced author embeddings $\mathbf{H}_u^{rf}$ of \textit{u} with $\mathbf{H}_u^{s}$, as well as the reader propagation roles' semantic matrix $\mathbf{H}^{role}$, as illustrated in Eq.\eqref{Eq_linear}-\eqref{Eq_fusion}.
{\setlength\abovedisplayskip{1pt}	
	\setlength\belowdisplayskip{1pt}
\begin{equation}\label{Eq_linear}
	\begin{aligned}
		\mathbf{M}_u^{s}&=\mathbf{W}_s\mathbf{H}_u^{s} + \mathbf{b}_s\\
		\mathbf{M}_u^{rf}&=\mathbf{W}_r\mathbf{H}_u^{rf} + \mathbf{b}_r\\
		\mathbf{M}^{role}&=\mathbf{W}_e\mathbf{H}^{role} + \mathbf{b}_e
	\end{aligned}
\end{equation} 
\begin{equation}\label{Eq_matt}
	\begin{aligned}
		\mathbf{a}_u^s &= \text{MATT}(\mathbf{M}_u^{rf}, \mathbf{M}_u^{s}, \mathbf{M}_u^{s})\\
		\mathbf{a}_u^{role} &= \text{MATT}(\mathbf{M}_u^{rf}, \mathbf{M}^{role}, \mathbf{M}^{role})
	\end{aligned}
\end{equation} 
\begin{equation}\label{Eq_gate}
	\begin{aligned}
		\mathbf{g}_u^s &= \mathbf{a}_u^s\odot\sigma(\mathbf{W}_{gs}\mathbf{a}_u^s + \mathbf{b}_{gs})\\
		\mathbf{g}_u^{role} &= \mathbf{a}_u^{role}\odot\sigma(\mathbf{W}_{gr}\mathbf{a}_u^{role} + \mathbf{b}_{gr})
	\end{aligned}
\end{equation} 
\begin{equation}\label{Eq_fusion}
	\mathbf{g}_u^{o}=\mathbf{g}_u^s\oplus\mathbf{g}_u^{role}\oplus(\mathbf{g}_u^s+\mathbf{g}_u^{role})\oplus(\mathbf{g}_u^s\odot\mathbf{g}_u^{role})
\end{equation} 
}where $\mathbf{W}_s\in \mathbb{R}^{N\times d_2}$, $\mathbf{W}_r\in \mathbb{R}^{3\times d_2}$, $\mathbf{W}_e\in \mathbb{R}^{4\times d_2}$, $\mathbf{W}_{gs,gr}\in \mathbb{R}^{d_2\times d_2}$, $\mathbf{b}_{s,r,e,gs,gr}\in \mathbb{R}^{d_2}$ are the trainable parameters, $\sigma$ is the sigmoid function, and $\odot$ is the element-wise multiplication.

$\mathbf{g}_u^{o}$ is then fed into the prediction layer to obtain the implicit emotion probability distribution. The loss is measured using the cross-entropy function, defined as:
{\setlength\abovedisplayskip{1pt}	
	\setlength\belowdisplayskip{1pt}
\begin{equation}\label{Eq_predict}
	\mathbf{\tilde{y}}=softmax(\mathbf{W}_o\mathbf{g}_u^{o} +\mathbf{b}_o)
\end{equation} 
\begin{equation}\label{Eq_loss}
	\mathfrak{L}=-\frac{1}{|D|}\sum_{\langle s, u \rangle\in D}\mathbf{y}_{\langle s, u \rangle}\log \mathbf{\tilde{y}}_{\langle s, u \rangle} + R(\theta)
\end{equation} 
}
where $\mathbf{W}_o\in \mathbb{R}^{l\times 4d_2}$, $\mathbf{b}_o\in \mathbb{R}^l$ are the weight and bias, respectively, $|D|$ is the size of dataset \textit{D}, $\mathbf{y}_{\langle s, u \rangle}$ is the golden label and $R(\theta)$ is the regularization.

\section{Experiments and Analysis}
\subsection{Dataset and Evaluation Metrics}
We construct two large-scale PIEA datasets that cover the mainstream Chinese and English social media platforms, Weibo and Twitter. The datasets consist of 7 implicit emotional categories, along with the corresponding user metadata including attributes, historical posts, and following relationships (only in Weibo). For details on the construction of the datasets and statistical analysis, please refer to Appendix~\ref*{Corpora}.

To evaluate the performance of the models, we utilize the F1 score for individual emotions, along with the overall  macro-F1 and accuracy metrics.

\subsection{Implementation Details}
Our framework is built upon PyTorch, leveraging P-Tuning v2 \cite{ptuning} for the fine-tuning of the reader agent, optimized by Adam on 2 RTX 3090 GPUs. The learning rates are [2e-6, 5e-6] for Weibo and Twitter datasets, respectively. The unified maximum sequence length \textit{N} is 128, and the batch size is 8. The maximum iterations \textit{K} for interactive propagation is limited to 1, which will be further detail discussed in section~\ref{iteration}. The GLM4 \footnote{\url{https://bigmodel.cn/}} and Qwen2.5-turbo \footnote{\url{https://qwenlm.github.io/zh/blog/qwen2.5-turbo/}} are adopted as the foundation LLMs for reader agents, respectively. The LLM-based encoders include ChatGLM-6B \cite{glm6b} and Qwen2.5-14B \cite{qwen2.5}. The embedding dimensions $[d_1, d_2]$ are configured as $[4096, 2048]$ (ChatGLM-6B) and $[5120, 2560]$ (Qwen2.5-14B). The hyperparameter top-\textit{k} in section~\ref{sec:global} in Weibo and Twitter is set to $[100, 100]$ (ChatGLM-based) and $[50, 100]$ (Qwen-based), respectively (detail discussed in section~\ref{readerscale}).

\subsection{Baselines}
To validate the advance of our model, we compare it with a range of SOTA sentiment/emotion analysis baselines, which can be divided into the following three categories.
\textbf{(1) Methods based on knowledge enhancement}: \textbf{KGAN} \cite{zhong2023knowledge} and \textbf{MMML} \cite{MMML24}.
\textbf{(2) Methods based on GNNs}: \textbf{RDGCN} \cite{zhao2024rdgcn} and \textbf{DAGCN} \cite{wang2024dagcn}.
\textbf{(3) Methods based on LLMs}: 
\textbf{GLM4}(online) \cite{glm2024chatglm}, \textbf{ChatGLM-6B}(ft) \cite{glm6b}, and \textbf{Qwen-14B}(ft) \cite{qwen2.5} are employed as the basic LLM baselines, (ft) means the LLM is fine-tuned with the training set. Then two sophisticated chain-of-thought-based semantic reasoning mechanisms, \textbf{THOR} \cite{fei2023} and \textbf{TOC} \cite{weinzierl-harabagiu-2024-tree} are combined with the basic LLMs. We also compared the latest \textbf{DeepSeek-R1} 671B \cite{deepseekr1}, optimized for reasoning tasks, to validate the advancement of our method.
See Appendix~\ref*{Baselines} for detailed descriptions.

\subsection{Overall Results}
The overall experimental results are displayed in Table~\ref{Tab:results}, the values in bold represent the best performance and underlined results are sub-optimal. We note that the responses of some LLM-based baselines may not strictly adhere to the required emotion categories or may fail to provide an answer, thus we treat these instances as a new error category. The results in parentheses represent the Macro-F1 score excluding this error category.

\begin{table*}[htb]
	\centering
	\small
	\setlength{\tabcolsep}{1.0mm}
	\caption{Overall comparison on the experimental datasets.}
	\label{Tab:results}
	\begin{tabular}{clccccccccc}
		\hline
		\multicolumn{1}{l}{Datasets} & Models & Happy  & Anger  & Sad    & Disgust& Fear   & Surprise  & Neutral& Macro-F1  & Accuracy  \\\hline
	&KGAN & 0.475 & 0.544    & 0.455 & 0.267 & 0.273 & 0.312 & 0.665  & 0.427 & 0.551\\
	&MMML & \underline{0.649}            & 0.507 & 0.466 & 0.160 & 0.230 & 0.360 & \textbf{0.690} & 0.438 & 0.595     \\\cline{2-11}
	&RDGCN& 0.512 & 0.275 & 0.329 & 0.113 & 0.291 & 0.000 & 0.650 & 0.310 & 0.515  \\
	&DAGCN          & 0.613 & 0.534 & 0.471 & 0.172 & 0.177 & 0.286 & 0.682 & 0.419 & 0.583  \\\cline{2-11}
	&GLM4 & 0.589 & 0.538 & 0.375 & 0.125 & 0.214 & 0.196 & 0.337 &  0.297(0.339) & 0.417  \\
	&THOR+GLM4            & 0.602 & 0.470 & 0.400 & 0.157 & 0.148 & 0.269 & 0.314 & 0.295(0.337) & 0.387  \\
	&TOC+GLM4             & 0.598 & 0.451 & 0.464 & 0.164 & 0.214 & 0.288 & 0.533 & 0.339(0.387) & 0.502  \\
	&ChatGLM-6B(ft)       & 0.505 & 0.323 & 0.462 & 0.221 & 0.267 & 0.125 & 0.637 & 0.282(0.363) & 0.501 \\	
	&THOR+ChatGLM-6B(ft)  & 0.627 & 0.422 & 0.476    & 0.227 & 0.300    & 0.292 & 0.598 & 0.420 & 0.529  \\
	&TOC+ChatGLM-6B(ft)   & 0.634 & 0.477 & 0.463 & 0.308 & 0.294 & 0.330 & 0.586 & 0.442 & 0.542    \\
	&Qwen-14B(ft)         & 0.611 & 0.454 & 0.397 & 0.078 & 0.133 & 0.103 & 0.485 & 0.323 & 0.463  \\
	&THOR+Qwen-14B(ft)        & 0.630 & 0.355 & 0.394 & 0.038 & 0.116 & 0.225 & 0.501 & 0.282(0.323) & 0.456  \\
	&TOC+Qwen-14B(ft)         & 0.609 & 0.480 & 0.441 & 0.175 & 0.167 & 0.355 & 0.674 & 0.414 & 0.574  \\
	& DeepSeek-R1 &\textbf{0.659} &	0.511& 	\underline{0.492} &	0.306& 	0.294 &	0.258 &	0.604 &	0.390(0.478) &	0.552 	\\\cline{2-11}
	&\textbf{ours RAPPIE(ChatGLM)} & 0.648               & \textbf{0.587} & \textbf{0.525} & \underline{0.344} & \textbf{0.353} & \textbf{0.485} & \underline{0.688}    & \textbf{0.519} & \textbf{0.617} \\
	\multirow{-11}{*}{Weibo} &\textbf{ours RAPPIE(Qwen)}    & 0.626 &	\underline{0.571}&	0.481&	\textbf{0.364}&	\underline{0.316}&	\underline{0.457}&	0.685&	\underline{0.500} 	&\underline{0.608}
	\\ 
	\hline
	\hline
	&KGAN                & 0.546 & 0.270 & 0.420 & 0.286 & 0.310 & 0.207 & 0.240  & 0.326  & 0.375      \\
	&MMML                & 0.623 & 0.266 & 0.533 & 0.354 & 0.286 & 0.318 & 0.293  & 0.382  & 0.474    \\\cline{2-11}
	&RDGCN               & 0.603 & 0.474 & 0.188 & 0.321 & 0.169 & 0.160 & 0.212  & 0.304   & 0.438     \\
	&DAGCN               & 0.618 & 0.269 & 0.512 & 0.326 & 0.283 & 0.318 & 0.264  & 0.370  & 0.460 \\\cline{2-11}
	&GLM4                & 0.700 & 0.360 & 0.579 & 0.345 & 0.341 & 0.363 & 0.227 & 0.364(0.416) & 0.480 \\
	&THOR+GLM4 &	\underline{0.724} &	0.344 &	0.567 &	\underline{0.437} &	0.362 &	0.389 &	0.086 &	0.415 &	\underline{0.513} \\
	& TOC+GLM4 &	\textbf{0.729} &	0.364 &	0.509 &	0.278 &	0.341 &	0.321 &	0.205 &	0.392 &	0.476 \\
	& ChatGLM-6B(ft)      & 0.227 & 0.242 & 0.262 & 0.309 & 0.030 & 0.135 & 0.271 & 0.185(0.211) & 0.254 \\
	& THOR+ChatGLM-6B(ft) & 0.657 & 0.171 & 0.360 & 0.414 & 0.330 & 0.317 & 0.286 & 0.362 & 0.434         \\
	& TOC+ChatGLM-6B(ft)  & 0.706 & 0.360 & 0.467 & 0.291 & 0.344 & 0.321 & 0.213 & 0.338(0.386) & 0.449  \\
	& Qwen-14B(ft)        & 0.703 & 0.339 & 0.568 & 0.401 & 0.116 & 0.299 & 0.267 & 0.385 & 0.480         \\
	& THOR+Qwen-14B(ft) &	0.708 &	0.345 &	\underline{0.647} &	0.206 &	0.397 &	\underline{0.426}&	0.191  &	0.365(0.417) &	0.491\\
	& TOC+Qwen-14B(ft) &	0.615 &	0.374 &	0.558 &	0.323 &	\textbf{0.452} &	0.302 &	0.317  &	0.368(0.420) &	0.446\\ 
	& DeepSeek-R1 &0.716 	&\underline{0.378}	&0.646 &	0.334 &	\underline{0.451} &	\textbf{0.429} &	0.228  &	0.398(0.455) &	0.499\\
	\cline{2-11}
	& \textbf{ours RAPPIE(ChatGLM)} &	0.636 &	0.321 &	0.568 &	\textbf{0.461} &	0.299 &	0.390 &	\underline{0.329} & \underline{0.429} &	0.490  \\
	\multirow{-11}{*}{Twitter}     & \textbf{ours RAPPIE(Qwen)} &	0.687 &	0.333 &	\textbf{0.657} &	0.394 &	0.374 &	0.271 &	\textbf{0.381}  &	\textbf{0.442} &	\textbf{0.527}
	\\
	\hline
	\end{tabular}
\end{table*}

From the data presented in Table~\ref{Tab:results}, on the Weibo dataset, our method achieved the best or sub-optimal results across all emotion categories except \textit{Happy}, and on Twitter, it also reached the best performance in multiple categories such as \textit{Sad}, \textit{Disgust}, and \textit{Neutral}.
Overall accuracy and macro-F1 on both datasets compared to the best baselines achieved 3.7\% (MMML), 17.4\% (TOC+ChatGLM-6B(ft)), 2.7\% (THOR+GLM4), and 6.5\% (THOR+GLM4) improvements, respectively.

Methods such as KGAN, and MMML primarily enhance model performance through external knowledge enhancement. However, these methods focus solely on knowledge related to the implicit emotional contents, lacking the modeling and integration of user-specific information. RDGCN and DAGCN are typical methods that employ GNNs to capture textual dependencies and sentiment aspects. Nevertheless, The lack of explicit emotional cues makes it challenging for syntactic dependency relations to achieve an effect similar to that observed in explicit emotional analysis. In comparison, our RAPPIE model incorporates user-content-reader feedback interactions, enhancing literal semantics through reader agents and a role-aware graph learning method that captures behaviorally-driven semantic information inspired by content.

Considering that LLM-based baselines, while they have robust semantic understanding capabilities, their standalone performance is limited. Chain-of-reasoning mechanism like THOR or TOC contributes to the semantic inference of implicit emotions, particularly when combined with fine-tuned LLMs. However, these methods primarily focus on the reasoning process of implicit emotional semantics, failing to model the outcomes of reasoning in human thinking from a behavioral perspective. The instability of some LLM-generated results also means that relying solely on LLMs is not the optimal solution for specific downstream tasks. In contrast, our approach surpasses the base LLMs and their reasoning-enhanced models, as well as the GLM4 or DeepSeek-R1 with tens of billions parameters. This demonstrates that agent-based user simulation introduces richer information about readers' feedback and interactive propagation beyond text semantics, which cannot be learned simply by increasing data or parameter scale. The DeepSeek-R1 model, enhanced by reinforcement learning-based inference, demonstrates superior fundamental analytical capabilities compared to GLM4 and Qwen. Despite its current limitation in incorporating information beyond semantics, it exhibits significant potential as a novel foundational LLM in the future.

Additionally, the model's performance on the Twitter dataset is generally inferior to that on the Weibo dataset, likely due to the absence of user follow relationship data, which diminishes the accuracy of reader propagation modeling. The differences in results between the two datasets indicate that PIEA in social media across diverse linguistic and cultural backgrounds is susceptible to foundation LLM. ChatGLM excels in simulating users with Chinese cultural backgrounds, while Qwen outperforms on more international social media.

\subsection{Ablation Study}
Table~\ref{Tab:ablation} summarizes the ablation study results. 
The exclusion of reader feedback from our fusion mechanism (\textit{w}/\textit{o} $\mathbf{M}_u^{rf}$) leads to a maximum performance degradation in macro-F1 of 8.5\% and 3.6\% on two respective datasets. The removal of propagation role information in the fusion layer (\textit{w}/\textit{o} $\mathbf{M}^{role}$) also impacts performance. During the propagation phase, the isolation of propagation role embeddings $\mathbf{H}^{role}$ on nodes intensifies user data sparsity, negatively affecting results. Models without comprehensive reader data, such as those with text + user attributes ($\mathbf{M}_u^{s}+\mathbf{H}_u^{at}$) or text alone ($\mathbf{M}_u^{s}$), show inferior performance due to the absence of critical reader information.

Our multi-view interactive learning reveals that each interaction behavior contributes to capturing reader feedback. Notably, the view of \textit{reposting with a comment} ($\mathcal{V}_c$) is most influential due to its comprehensive behavioral and emotional information. In contrast, simple \textit{reposting} ($\mathcal{V}_r$) behaviors often do not reflect readers' emotional feedback directly. Due to the inherent incompleteness of following relationships and the broadcast approach used in simulating user feedback, removing view $\mathcal{V}_f$ has a minimal impact on overall performance. 

As shown in our Twitter experiments (without $\mathcal{V}_f$), the model remains effective even when following relations are missing. Our model simulates reader feedback propagation via a broadcast mechanism across relevant readers with shared interests (rather than relying on explicit user following relations). This can ensure that each user receives feedback from a reader group of the same size, eliminating bias that might arise from differences in reader scale. In addition, the <user, post> pairs with low-quality feedback naturally become minority edges in the broadcast network, minimally affecting multi-view interaction learning. Unlike static follower graphs, broadcasting mimics the recommendation mechanism of trending topics on social media platforms, uncovering latent user interactions.
\begin{table}[!h]
	\centering
	\small
	\setlength{\tabcolsep}{1.5mm}
	\caption{Performance of ablation experiments.}
	\label{Tab:ablation}
	\begin{tabular}{lcccc}
		\hline
		& \multicolumn{2}{c}{Weibo(Macro-F1)}  & \multicolumn{2}{c}{Twitter(Macro-F1)}      \\ 	\cline{2-5}
		Models	& ChatGLM  & Qwen  & ChatGLM  & Qwen  \\ \hline
		\textbf{RAPPIE}   &  \textbf{0.519} &  \textbf{0.500} &  \textbf{0.429} &  \textbf{0.442} \\ \hline
		\textit{w}/\textit{o} $\mathbf{M}_u^{rf}$ & 0.454  & 0.415  & 0.393  & 0.417  \\
		\textit{w}/\textit{o} $\mathbf{M}^{role}$ & 0.491  & 0.438  & 0.399  & 0.425  \\		
		\textit{w}/\textit{o} $\mathbf{H}^{role}$  & 0.472  & 0.424  & 0.390  & 0.402  \\	
		$\mathbf{M}_u^{s}+\mathbf{H}_u^{at}$   & 0.435  & 0.400  & 0.358  & 0.390  \\	
		$\mathbf{M}_u^{s}$ \textit{only}       & 0.433  & 0.395  & 0.306  & 0.367 \\
		\hline
		\textit{w}/\textit{o} $\mathcal{V}_r$       & 0.474  & 0.445  & 0.397  & 0.424  \\
		\textit{w}/\textit{o} $\mathcal{V}_c$       & 0.459  & 0.439  & 0.396  & 0.422  \\ 
		\textit{w}/\textit{o} $\mathcal{V}_f$      & 0.484  & 0.465  & -  & -  \\\hline
	\end{tabular}
\end{table}
\subsection{Influence of Interactive Propagation}
\label{iteration}
\begin{figure}[!h]
	\centering
	\includegraphics[width=0.99\linewidth]{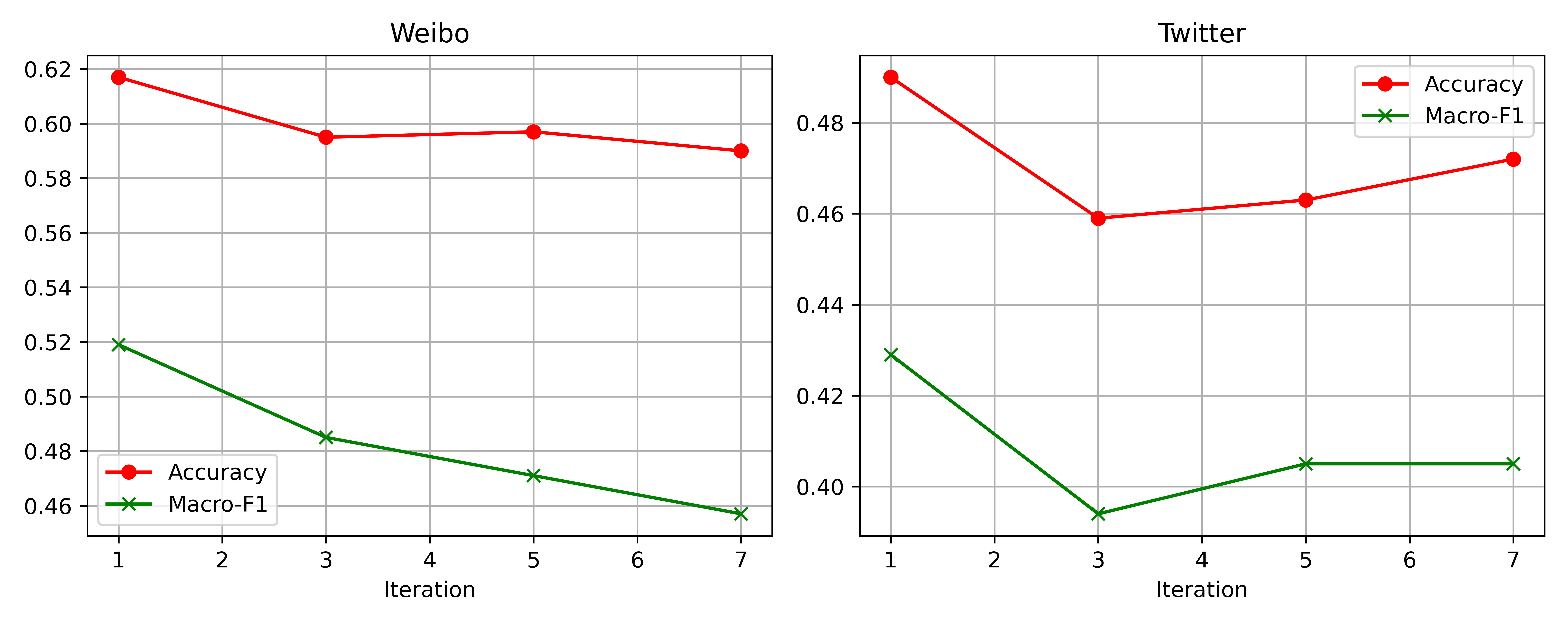} 
	\caption{Impact analysis of the maximum iteration in multi-view interactive propagation graph learning.}
	\label{fig_iteration}
\end{figure}

In this section, we investigate the impact of the maximum iterations \textit{K} for interactive propagation with RAPPIE (ChatGLM). As illustrated in Figure~\ref{fig_iteration}, excessive propagation rounds do not yield positive performance gains. This may be attributed to: (1) the tendency of human readers to generate feedback primarily upon initial exposure to content and other readers' comments, maintaining their stance until new information is received. (2) the over-smoothing issue\cite{wu2023demystifying} inherent in GNNs, where excessive iterations lead to indistinguishable node representations.

\subsection{Influence of Reader Scales}
\label{readerscale}
\begin{figure}[!h]
	\centering
	\includegraphics[width=0.99\linewidth]{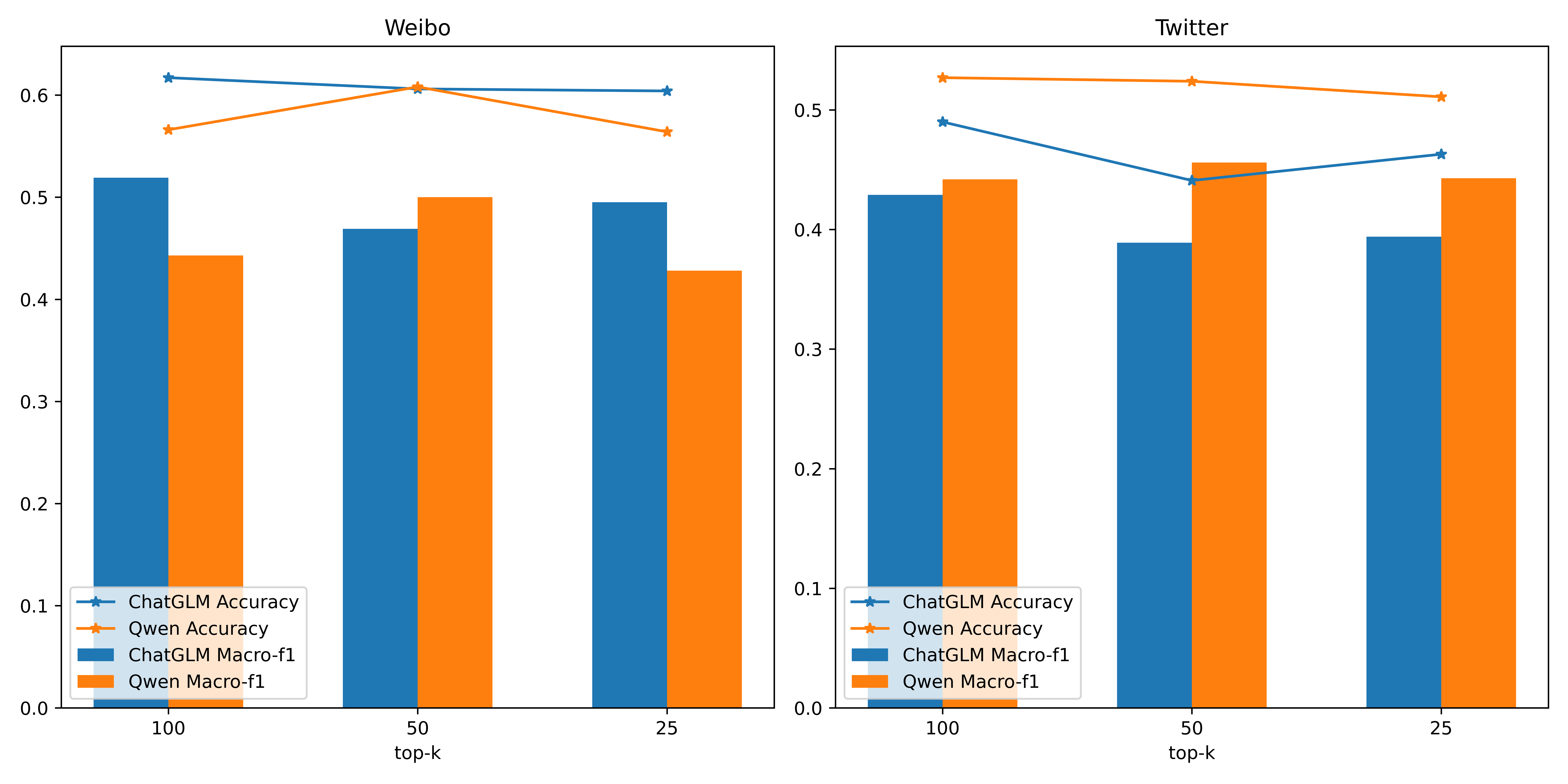} 
	\caption{Influence of different reader scales.}
	\label{fig_k}
\end{figure}

The scale of relevant readers is primarily influenced by the hyperparameter $k$ in $p_r^{top-k}$ mentioned in section~\ref{sec:global}, which is designed to: (1) broadcasts post to readers with aligned interests (reducing noise from unrelated users); (2) balances computational cost and signal diversity. We have set $k=[100,50,25]$ for detailed analysis, and the statistics of reader user scale and agent simulation behavior are shown in Table~\ref{tab:readerback} of Appendix~\ref{readerback}. The impact of different reader scales on the prediction results is illustrated in Figure~\ref{fig_k}. 

As the hyperparameter \textit{k} decreases, the model broadcasts to a smaller scope of reader groups to obtain feedback. Although smaller reader groups have interests more similar to the content author, this leads to a loss of generality in reader feedback, falling into a situation akin to an ``information cocoon.'' Therefore, an adequate scale of reader groups is necessary to obtain comprehensive and complete reader feedback. 

\subsection{Case Study}
We conduct a detailed case analysis to provide a clearer intuitive impression of our approach. Due to the transparency of the reasoning process in the LLM-based method, we compared our method with THOR+GLM4, THOR+Qwen-14B(ft), TOC+GLM4, and TOC+Qwen-14B(ft) in recognizing implicit emotions for the given content and its associated historical posts list, illustrated in Figures~\ref{fig_case1}-\ref{fig_case3} in Appendix~\ref*{detailcase}. It is evident that while LLMs + chain-of-reasoning mechanisms can effectively identify emotional triggers within the author's content, their reasoning tends to be grounded in rational and logical analysis. 
In contrast, the reader agent in RAPPIE demonstrates a superior ability to provide emotional interpretations from an empathetic perspective. Consequently, its predictions are more congruent with the author's genuine emotions.

\section{Conclusion}
In this paper, we introduce the RAPPIE model, which targets the PIEA task by addressing the challenges in reader feedback acquisition and interaction modeling. RAPPIE leverages LLM-based reader agents and a role-aware multi-view interactive propagation graph to facilitate personalized emotion modeling. Additionally, we present two new PIEA datasets covering different linguistic and cultural backgrounds, enriched with detailed user metadata annotations. Our experiments on these datasets demonstrate the RAPPIE model’s efficacy, outperforming best SOTA baseline with improvements of 17.4\% and 6.5\% in macro-F1. In the future, we will focus on the group-level analysis of implicit emotions in terms of detection, propagation, and drifting.

\section*{Limitations}
Our study did not explicitly address the potential biases inherent in LLMs during the simulation of reader behaviors. This omission is based on existing agent-based research in areas like fake news detection and dialogue, which suggests that in most circumstances LLM biases are not necessarily determinative for general downstream tasks. Our model primarily leverages user-generated content and user-attribute data. During dataset construction, we reviewed and excluded content that pertains to race, nationality, or other obvious factors prone to evoke discrimination or bias. Nevertheless, it is not possible to guarantee the absence of bias in the agent-based reader simulation process. We conducted experiments using two mainstream open-source LLMs across Chinese and English social media datasets, reflecting specific model biases to some extent. In future work, addressing the mitigation of potential implicit biases in LLMs during the construction of reader agents, in light of the diverse linguistic and cultural backgrounds of users, remains a crucial research task.

\section*{Ethics Statement}
In the process of generating comments by the reader agent, no explicit ethical or bias constraints were applied, relying primarily on the built-in ethical review mechanisms of the online LLM service providers. Thus, it is crucial to note that the data may contain potentially misleading content. As a result, manual verification is deemed necessary when applying these comments to real-world practices.

\section*{Acknowledgements}
The works described in this paper are supported by the National Natural Science Foundation of China (U24A20335, 62476162, 62376143, 62272286, 62473241), the Natural Science Foundation of Shanxi Province, China (202303021211021). Thanks to Bayou Tech (Hong Kong) Limited for providing the data used in this paper free of charge.

\bibliography{acl_latex}

\begin{thebibliography}{36}
\providecommand{\natexlab}[1]{#1}

\bibitem[{Aher et~al.(2023)Aher, Arriaga, and Kalai}]{pmlr-v202-aher23a}
Gati~V Aher, Rosa~I. Arriaga, and Adam~Tauman Kalai. 2023.
\newblock Using large language models to simulate multiple humans and replicate
  human subject studies.
\newblock In \emph{Proceedings of the 40th International Conference on Machine
  Learning}, volume 202 of \emph{Proceedings of Machine Learning Research},
  pages 337--371. PMLR.

\bibitem[{Biddle(2013)}]{biddle2013role}
Bruce~J Biddle. 2013.
\newblock \emph{Role theory: Expectations, identities, and behaviors}.
\newblock Academic press.

\bibitem[{Cai et~al.(2021)Cai, Xia, and Yu}]{cai2021}
Hongjie Cai, Rui Xia, and Jianfei Yu. 2021.
\newblock \href {https://doi.org/10.18653/v1/2021.acl-long.29}
  {Aspect-category-opinion-sentiment quadruple extraction with implicit aspects
  and opinions}.
\newblock In \emph{Proceedings of the 59th Annual Meeting of the Association
  for Computational Linguistics and the 11th International Joint Conference on
  Natural Language Processing (Volume 1: Long Papers)}, pages 340--350, Online.
  Association for Computational Linguistics.

\bibitem[{Davidson et~al.(2023)Davidson, Romeo, Shu, Gung, Gupta, Mansour, and
  Zhang}]{davidson2023user}
Sam Davidson, Salvatore Romeo, Raphael Shu, James Gung, Arshit Gupta, Saab
  Mansour, and Yi~Zhang. 2023.
\newblock User simulation with large language models for evaluating
  task-oriented dialogue.
\newblock \emph{arXiv preprint arXiv:2309.13233}.

\bibitem[{DeepSeek-AI et~al.(2025)DeepSeek-AI, Guo, Yang, and
  et~al.}]{deepseekr1}
DeepSeek-AI, Daya Guo, Dejian Yang, and et~al. 2025.
\newblock \href {https://arxiv.org/abs/2501.12948} {Deepseek-r1: Incentivizing
  reasoning capability in llms via reinforcement learning}.
\newblock \emph{Preprint}, arXiv:2501.12948.

\bibitem[{Du et~al.(2022)Du, Qian, Liu, Ding, Qiu, Yang, and Tang}]{glm6b}
Zhengxiao Du, Yujie Qian, Xiao Liu, Ming Ding, Jiezhong Qiu, Zhilin Yang, and
  Jie Tang. 2022.
\newblock {GLM}: General language model pretraining with autoregressive blank
  infilling.
\newblock In \emph{Proceedings of the 60th Annual Meeting of the Association
  for Computational Linguistics (Volume 1: Long Papers)}, pages 320--335,
  Dublin, Ireland. Association for Computational Linguistics.

\bibitem[{Fei et~al.(2023)Fei, Li, Liu, Bing, Li, and Chua}]{fei2023}
Hao Fei, Bobo Li, Qian Liu, Lidong Bing, Fei Li, and Tat-Seng Chua. 2023.
\newblock \href {https://doi.org/10.18653/v1/2023.acl-short.101} {Reasoning
  implicit sentiment with chain-of-thought prompting}.
\newblock In \emph{Proceedings of the 61st Annual Meeting of the Association
  for Computational Linguistics (Volume 2: Short Papers)}, pages 1171--1182,
  Toronto, Canada. Association for Computational Linguistics.

\bibitem[{GLM et~al.(2024)GLM, Zeng, Xu, and et~al.}]{glm2024chatglm}
Team GLM, Aohan Zeng, Bin Xu, and et~al. 2024.
\newblock Chatglm: A family of large language models from glm-130b to glm-4 all
  tools.
\newblock \emph{arXiv preprint arXiv:2406.12793}.

\bibitem[{Guo et~al.(2024)Guo, Cheng, Liang, Chen, and
  Han}]{guo2024integrating}
Naicheng Guo, Hongwei Cheng, Qianqiao Liang, Linxun Chen, and Bing Han. 2024.
\newblock Integrating large language models with graphical session-based
  recommendation.
\newblock \emph{arXiv preprint arXiv:2402.16539}.

\bibitem[{He et~al.(2020)He, Deng, Wang, Li, Zhang, and Wang}]{he2020lightgcn}
Xiangnan He, Kuan Deng, Xiang Wang, Yan Li, Yongdong Zhang, and Meng Wang.
  2020.
\newblock Lightgcn: Simplifying and powering graph convolution network for
  recommendation.
\newblock In \emph{Proceedings of the 43rd International ACM SIGIR conference
  on research and development in Information Retrieval}, pages 639--648.

\bibitem[{Huang et~al.(2024)Huang, Yang, Jiang, Bei, Zhang, and
  Chen}]{huang2024large}
Feiran Huang, Zhenghang Yang, Junyi Jiang, Yuanchen Bei, Yijie Zhang, and Hao
  Chen. 2024.
\newblock Large language model interaction simulator for cold-start item
  recommendation.
\newblock \emph{arXiv preprint arXiv:2402.09176}.

\bibitem[{Kertkeidkachorn and Shirai(2023)}]{kertkeidkachorn2023}
Natthawut Kertkeidkachorn and Kiyoaki Shirai. 2023.
\newblock \href {https://doi.org/10.18653/v1/2023.findings-acl.547} {Sentiment
  analysis using the relationship between users and products}.
\newblock In \emph{Findings of the Association for Computational Linguistics:
  ACL 2023}, pages 8611--8618, Toronto, Canada. Association for Computational
  Linguistics.

\bibitem[{Li et~al.(2024)Li, Zhang, Du, Pang, Liu, Guo, Shen, and
  Liu}]{li2024your}
Tianlin Li, Xiaoyu Zhang, Chao Du, Tianyu Pang, Qian Liu, Qing Guo, Chao Shen,
  and Yang Liu. 2024.
\newblock Your large language model is secretly a fairness proponent and you
  should prompt it like one.
\newblock \emph{arXiv preprint arXiv:2402.12150}.

\bibitem[{Li et~al.(2021)Li, Zou, Zhang, Zhang, and Wei}]{li2021}
Zhengyan Li, Yicheng Zou, Chong Zhang, Qi~Zhang, and Zhongyu Wei. 2021.
\newblock \href {https://doi.org/10.18653/v1/2021.emnlp-main.22} {Learning
  implicit sentiment in aspect-based sentiment analysis with supervised
  contrastive pre-training}.
\newblock In \emph{Proceedings of the 2021 Conference on Empirical Methods in
  Natural Language Processing}, pages 246--256, Online and Punta Cana,
  Dominican Republic. Association for Computational Linguistics.

\bibitem[{Liang et~al.(2024)Liang, Huang, Wang, and Yu}]{multiview}
Youwei Liang, Dong Huang, Chang-Dong Wang, and Philip~S. Yu. 2024.
\newblock Multi-view graph learning by joint modeling of consistency and
  inconsistency.
\newblock \emph{IEEE Transactions on Neural Networks and Learning Systems},
  35:2848--2862.

\bibitem[{Liao et~al.(2023)Liao, Lei, Wang, Zheng, and Han}]{liao2023}
Jian Liao, Jia Lei, Suge Wang, Jianxing Zheng, and Xiaoqing Han. 2023.
\newblock \href {https://doi.org/10.1109/CAC59555.2023.10451456} {Personalized
  implicit sentiment analysis based on multi-view fusion of implicit user
  preference}.
\newblock In \emph{2023 China Automation Congress (CAC)}, pages 6394--6399.

\bibitem[{Liao et~al.(2022)Liao, Wang, Chen, Wang, and Zhang}]{LIAO2022}
Jian Liao, Min Wang, Xin Chen, Suge Wang, and Kai Zhang. 2022.
\newblock Dynamic commonsense knowledge fused method for {C}hinese implicit
  sentiment analysis.
\newblock \emph{Information Processing \& Management}, 59(3):102934.

\bibitem[{Liu et~al.(2022)Liu, Ji, Fu, Tam, Du, Yang, and Tang}]{ptuning}
Xiao Liu, Kaixuan Ji, Yicheng Fu, Weng Tam, Zhengxiao Du, Zhilin Yang, and Jie
  Tang. 2022.
\newblock \href {https://doi.org/10.18653/v1/2022.acl-short.8} {{P}-tuning:
  Prompt tuning can be comparable to fine-tuning across scales and tasks}.
\newblock In \emph{Proceedings of the 60th Annual Meeting of the Association
  for Computational Linguistics (Volume 2: Short Papers)}, pages 61--68,
  Dublin, Ireland. Association for Computational Linguistics.

\bibitem[{Lyu et~al.(2023)Lyu, Yang, Zhang, Graham, and Foster}]{lyu2023}
Chenyang Lyu, Linyi Yang, Yue Zhang, Yvette Graham, and Jennifer Foster. 2023.
\newblock \href {https://doi.org/10.18653/v1/2023.findings-acl.92} {Exploiting
  rich textual user-product context for improving personalized sentiment
  analysis}.
\newblock In \emph{Findings of the Association for Computational Linguistics:
  ACL 2023}, pages 1419--1429, Toronto, Canada. Association for Computational
  Linguistics.

\bibitem[{Ma et~al.(2023)Ma, Hu, Liu, Yang, Li, Yu, and Wen}]{ma2023}
Fukun Ma, Xuming Hu, Aiwei Liu, Yawen Yang, Shuang Li, Philip~S. Yu, and Lijie
  Wen. 2023.
\newblock \href {https://doi.org/10.18653/v1/2023.acl-long.19} {{AMR}-based
  network for aspect-based sentiment analysis}.
\newblock In \emph{Proceedings of the 61st Annual Meeting of the Association
  for Computational Linguistics (Volume 1: Long Papers)}, pages 322--337,
  Toronto, Canada. Association for Computational Linguistics.

\bibitem[{Mireshghallah et~al.(2022)Mireshghallah, Shrivastava, Shokouhi,
  Berg-Kirkpatrick, Sim, and Dimitriadis}]{mireshghallah2022}
Fatemehsadat Mireshghallah, Vaishnavi Shrivastava, Milad Shokouhi, Taylor
  Berg-Kirkpatrick, Robert Sim, and Dimitrios Dimitriadis. 2022.
\newblock \href {https://doi.org/10.18653/v1/2022.naacl-main.252}
  {{U}ser{I}dentifier: Implicit user representations for simple and effective
  personalized sentiment analysis}.
\newblock In \emph{Proceedings of the 2022 Conference of the North American
  Chapter of the Association for Computational Linguistics: Human Language
  Technologies}, pages 3449--3456, Seattle, United States. Association for
  Computational Linguistics.

\bibitem[{Nan et~al.(2024)Nan, Sheng, Cao, Hu, Wang, and Li}]{nan2024let}
Qiong Nan, Qiang Sheng, Juan Cao, Beizhe Hu, Danding Wang, and Jintao Li. 2024.
\newblock Let silence speak: Enhancing fake news detection with generated
  comments from large language models.
\newblock \emph{arXiv preprint arXiv:2405.16631}.

\bibitem[{Owoicho et~al.(2023)Owoicho, Sekulic, Aliannejadi, Dalton, and
  Crestani}]{owoicho2023exploiting}
Paul Owoicho, Ivan Sekulic, Mohammad Aliannejadi, Jeffrey Dalton, and Fabio
  Crestani. 2023.
\newblock Exploiting simulated user feedback for conversational search:
  Ranking, rewriting, and beyond.
\newblock In \emph{Proceedings of the 46th International ACM SIGIR Conference
  on Research and Development in Information Retrieval}, pages 632--642.

\bibitem[{Schmid(2010)}]{Schmid2010}
Wolf Schmid. 2010.
\newblock \href {https://doi.org/doi:10.1515/9783110226324} {\emph{Narratology:
  An Introduction}}.
\newblock De Gruyter, Berlin, New York.

\bibitem[{Sohn(2022)}]{Sohn2022Spiral}
Dongyoung Sohn. 2022.
\newblock Spiral of silence in the social media era: A simulation approach to
  the interplay between social networks and mass media.
\newblock \emph{Communication Research}, 49(1):139--166.

\bibitem[{Team(2024)}]{qwen2.5}
Qwen Team. 2024.
\newblock \href {https://qwenlm.github.io/blog/qwen2.5/} {Qwen2.5: A party of
  foundation models}.

\bibitem[{Tran et~al.(2023)Tran, Shirai, and Kertkeidkachorn}]{tran2023}
Tu~Tran, Kiyoaki Shirai, and Natthawut Kertkeidkachorn. 2023.
\newblock \href {https://doi.org/10.18653/v1/2023.findings-acl.323} {Text
  generation model enhanced with semantic information in aspect category
  sentiment analysis}.
\newblock In \emph{Findings of the Association for Computational Linguistics:
  ACL 2023}, pages 5256--5268, Toronto, Canada. Association for Computational
  Linguistics.

\bibitem[{Wang et~al.(2022)Wang, Zhou, Sun, Ye, Gui, Zhang, and
  Huang}]{wang2022causal}
Siyin Wang, Jie Zhou, Changzhi Sun, Junjie Ye, Tao Gui, Qi~Zhang, and Xuanjing
  Huang. 2022.
\newblock Causal intervention improves implicit sentiment analysis.
\newblock In \emph{Proceedings of the 29th International Conference on
  Computational Linguistics}, pages 6966--6977.

\bibitem[{Wang et~al.(2024{\natexlab{a}})Wang, Wang, Cheng, Li, and
  Liu}]{wang2024deem}
Xiaolong Wang, Yile Wang, Sijie Cheng, Peng Li, and Yang Liu.
  2024{\natexlab{a}}.
\newblock \href {https://aclanthology.org/2024.lrec-main.405} {{DEEM}: Dynamic
  experienced expert modeling for stance detection}.
\newblock In \emph{Proceedings of the 2024 Joint International Conference on
  Computational Linguistics, Language Resources and Evaluation (LREC-COLING
  2024)}, pages 4530--4541, Torino, Italia. ELRA and ICCL.

\bibitem[{Wang et~al.(2024{\natexlab{b}})Wang, Zhang, Yang, Guo, and
  Li}]{wang2024dagcn}
Zhihao Wang, Bo~Zhang, Ru~Yang, Chang Guo, and Maozhen Li. 2024{\natexlab{b}}.
\newblock \href {https://doi.org/10.18653/v1/2024.findings-naacl.120} {{DAGCN}:
  Distance-based and aspect-oriented graph convolutional network for
  aspect-based sentiment analysis}.
\newblock In \emph{Findings of the Association for Computational Linguistics:
  NAACL 2024}, pages 1863--1876, Mexico City, Mexico. Association for
  Computational Linguistics.

\bibitem[{Weinzierl and Harabagiu(2024)}]{weinzierl-harabagiu-2024-tree}
Maxwell Weinzierl and Sanda Harabagiu. 2024.
\newblock \href {https://doi.org/10.18653/v1/2024.acl-long.49}
  {Tree-of-counterfactual prompting for zero-shot stance detection}.
\newblock In \emph{Proceedings of the 62nd Annual Meeting of the Association
  for Computational Linguistics (Volume 1: Long Papers)}, pages 861--880,
  Bangkok, Thailand. Association for Computational Linguistics.

\bibitem[{Wu et~al.(2023)Wu, Ajorlou, Wu, and Jadbabaie}]{wu2023demystifying}
Xinyi Wu, Amir Ajorlou, Zihui Wu, and Ali Jadbabaie. 2023.
\newblock Demystifying oversmoothing in attention-based graph neural networks.
\newblock \emph{Advances in Neural Information Processing Systems}, 36.

\bibitem[{Wu et~al.(2024)Wu, Gong, Koo, and Hirschberg}]{MMML24}
Zehui Wu, Ziwei Gong, Jaywon Koo, and Julia Hirschberg. 2024.
\newblock \href {https://doi.org/10.18653/v1/2024.naacl-long.197} {Multimodal
  multi-loss fusion network for sentiment analysis}.
\newblock In \emph{Proceedings of the 2024 Conference of the North American
  Chapter of the Association for Computational Linguistics: Human Language
  Technologies (Volume 1: Long Papers)}, pages 3588--3602, Mexico City, Mexico.
  Association for Computational Linguistics.

\bibitem[{Xie et~al.(2024)Xie, Chen, Jia, Ye, Shu, Bibi, Hu, Torr, Ghanem, and
  Li}]{xie2024can}
Chengxing Xie, Canyu Chen, Feiran Jia, Ziyu Ye, Kai Shu, Adel Bibi, Ziniu Hu,
  Philip Torr, Bernard Ghanem, and Guohao Li. 2024.
\newblock Can large language model agents simulate human trust behaviors?
\newblock \emph{arXiv preprint arXiv:2402.04559}.

\bibitem[{Zhao et~al.(2024)Zhao, Peng, Dai, Bai, Peng, Liu, Guo, and
  Yu}]{zhao2024rdgcn}
Xusheng Zhao, Hao Peng, Qiong Dai, Xu~Bai, Huailiang Peng, Yanbing Liu,
  Qinglang Guo, and Philip~S Yu. 2024.
\newblock Rdgcn: Reinforced dependency graph convolutional network for
  aspect-based sentiment analysis.
\newblock In \emph{Proceedings of the 17th ACM International Conference on Web
  Search and Data Mining}, pages 976--984.

\bibitem[{Zhong et~al.(2023)Zhong, Ding, Liu, Du, Jin, and
  Tao}]{zhong2023knowledge}
Qihuang Zhong, Liang Ding, Juhua Liu, Bo~Du, Hua Jin, and Dacheng Tao. 2023.
\newblock Knowledge graph augmented network towards multiview representation
  learning for aspect-based sentiment analysis.
\newblock \emph{IEEE Transactions on knowledge and data engineering},
  35(10):10098--10111.

\end{thebibliography}

\appendix
\section{Details Prompt for Reader Agent Construction}
\label{agentprompt}
\begin{figure}[ht]
	\centering
	\includegraphics[width=0.99\linewidth]{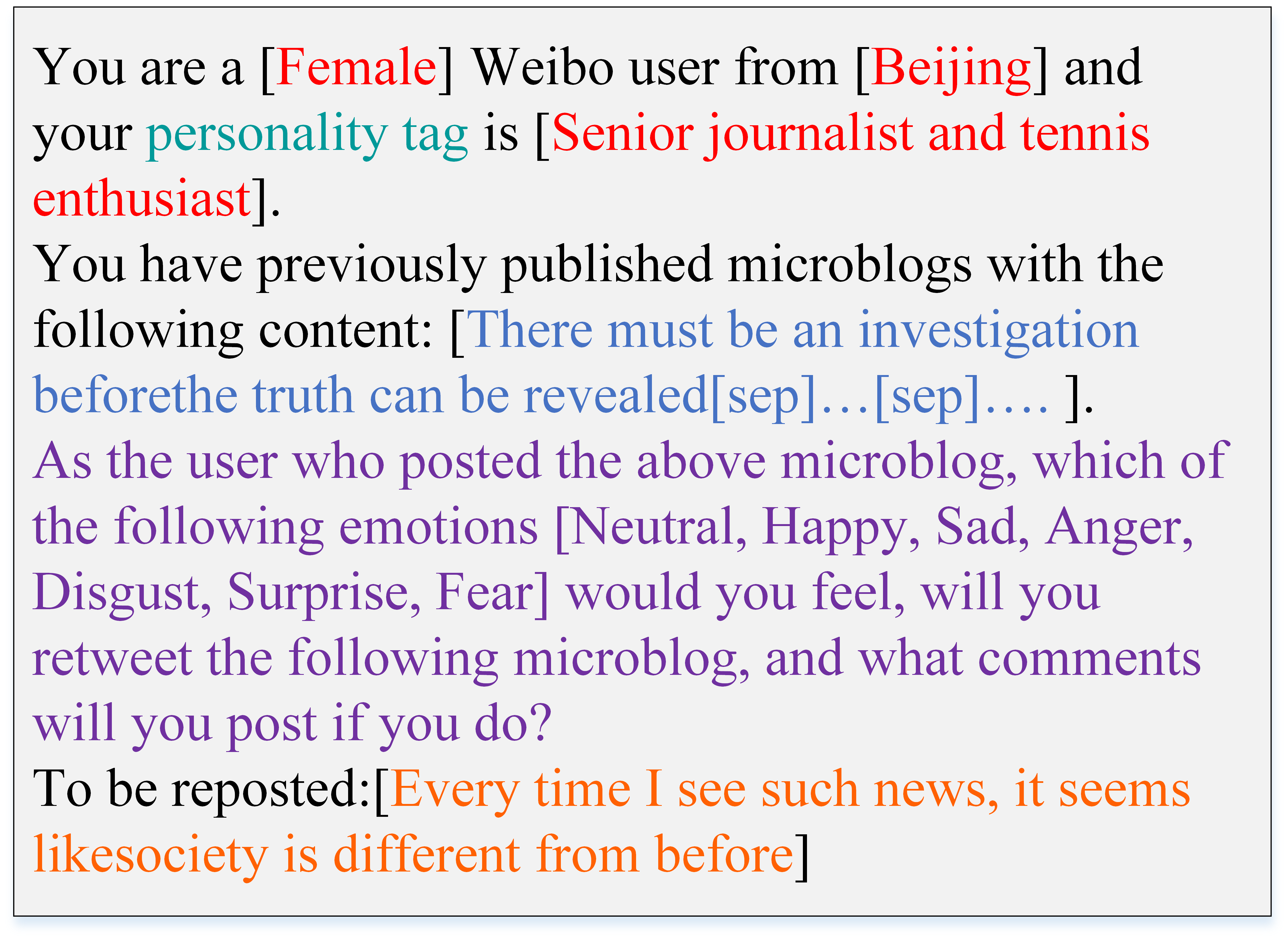} 
	\caption{An example of the prompt template for reader agent creation.}
	\label{fig_prompt_agent}
\end{figure}

Figure~\ref{fig_prompt_agent} shows an example of the prompt template for creating a reader agent. The red contents correspond to $u_a^g$,  $u_a^r$, and  $u_a^t$ of the reader attributes $u_a$. The blue content represents the reader's historical posts $u_h$, with multiple entries separated by [sep]. The orange part is the content to be reposted.

\section{Details of the Reader Feedback Simulation Process}
\label{readerback}
During the global multi-behavioral interactive overlapping network, we construct prompts based on $\langle p, r \rangle$ pairs, and input them into the reader \textit{r}'s agent. Due to the incompleteness of the user-following relationships, we adopt a broadcasting approach rather than relying on user-following connections for propagation simulation, to obtain more comprehensive simulation results.
To address the issue of propagation noise, we employ a text similarity calculation method based on \textit{Tf-Idf} to filter the $top-k$ posts semantically relevant to the reader \textit{r}'s historical posts. Detailed statistics of the reader feedback simulation results are shown in Table~\ref{tab:readerback}. The GLM4 and Qwen2.5-turbo (Qwen for short) are adopted as the foundation LLMs for reader agents, respectively.

\begin{table*}[htbp]
	\centering
	\small
	\setlength{\tabcolsep}{1.8mm}
	\caption{The statistics of the reader feedback simulation results on different reader scales.}
	\label{tab:readerback}
	\begin{tabular}{cccccc}
		\hline
		top-$k$ & Agent in datasets & Reposting & Reposting with a comment & No repost & Total  \\ \hline
		& GLM4 in Weibo & 97505     & 116947      & 136348    & 350800 \\
		&Qwen in Weibo & 63880     & 147430      & 139490    & 350800 \\
		&GLM4 in Twitter  & 54860     & 281779      & 230961    & 567600 \\
		\multirow{-4}{*}{100} &Qwen in Twitter & 151567    & 130808      & 285225    & 567600 \\\hline
		& GLM4 in Weibo & 46436 & 51562 & 77402 & 175400 \\
		&Qwen in Weibo & 30109 & 68913 & 76378 & 175400 \\
		&GLM4 in Twitter  & 23632 & 138279 & 121889 & 283800 \\
		\multirow{-4}{*}{50} &Qwen in Twitter & 70776 & 69244 & 143780 & 283800 \\\hline
		& GLM4 in Weibo & 23377 & 25647 & 38676 & 87700 \\
		&Qwen in Weibo & 14959 & 34599 & 38142 & 87700 \\
		&GLM4 in Twitter  & 12474 & 69303 & 60123 & 141900 \\
		\multirow{-4}{*}{25} &Qwen in Twitter & 34304 & 35102 & 72494 & 141900 \\
		\hline
	\end{tabular}%
\end{table*}

\section{Detailed explanation of the reader propagation roles system}
\label{proprole}
In this study, drawing upon the role behavior theory in communication, we establish a quaternary role system based on the dimensions of users' rationality-sensibility and action-hesitation and define four propagation roles: Emotional person, Gatekeeper, Onlooker, and Rationalist, detailed as the following Table~\ref{tab:role1}.
\begin{table*}[!ht]
	\centering
	\small
	\setlength{\tabcolsep}{1.0mm}
	\caption{Reader propagation roles system and their detailed descriptions.}
	\label{tab:role1}
	\begin{tabular}{p{0.14\textwidth}p{0.3\textwidth}p{0.5\textwidth}}
		\hline
		Propagation role & Description & Example  \\\hline
		Emotional person & An emotional person tends to be highly sensitive to certain types of emotions, readily experiencing and reacting to events that elicit such feelings. They possess a strong propensity to disseminate these events, often augmenting the associated emotions through their communications. & A person with a propensity for anger might react strongly to an incident like "a teacher in Canton physically punishing a student," not only spreading the news of the event but also accompanying their transmission with comments that reflect their anger. Similarly, the same individual might be sensitive to positive emotions and respond to an event like ``a woman being arrested for spreading rumors'' with a sense of satisfaction, thereby contributing to the propagation of both the event and the corresponding emotional response. \\ \hline
		Gatekeeper & The gatekeeper exhibits relative rationality towards a certain emotional type, participating in the dissemination of events while concurrently mitigating the corresponding emotion. & This individual maintains a rational perspective on anger and plays a role in weakening the anger associated with the source event by issuing objective and soothing statements or actively disseminating events such as ``medical professor questioning bloodstains'' within the propagation network. \\ \hline
		Rationalist & The rationalist exhibits relative rationality towards a certain emotional type, making it less prone to triggering that emotion in response to events. This role participates in the dissemination of the event without attaching the corresponding emotion. & In the case of anger, the individual remains relatively rational, merely contributing to the amplification of the event's prominence without embodying the anger. Over time, this role may shift to become an emotional person, a gatekeeper, or an onlooker. \\ \hline
		Onlooker & The onlooker maintain a passive observation stance towards events, generally not participating in the dissemination of information. & They typically function as information recipients, merely observing the development trend of events, seldom expressing personal opinions, and not engaging in retweets. However, in the context of certain events, they may transition into the role of a rationalist.\\ \hline
	\end{tabular}
\end{table*}

To capture the reader propagation roles' features, we concatenate each role's description with example content and construct a prompt as: Prompt=``\textit{The description of the propagation role is [Description], with an example [Example]. Explain the role [Propagation Role] based on the above description and example.}'' The filled prompt is then fed into the online LLM GLM4. The enriched explanation of each role is shown in Table~\ref{tab:role2}.
	
\begin{table*}[htbp]
	\centering
	\small
	\setlength{\tabcolsep}{1.0mm}
	\caption{The enriched explanation of each role through LLM.}
	\label{tab:role2}
		\begin{tabular}{p{0.14\textwidth}p{0.8\textwidth}}
			\hline
			Propagation Role & Enriched Explanation \\\hline
			Emotional person & The term ``Emotional Person'' refers to a group of individuals on social media platforms who exhibit a high level of sensitivity to specific events or topics and are prone to experiencing particular emotions upon receiving relevant information. They have a strong propensity to disseminate information and actively share these events, often accompanied by expressions of their own emotions in the process of forwarding. Based on the characteristics and examples provided, we can identify the following traits of Emotional Persons: Emotional Sensitivity: They have a heightened sensitivity to certain types of emotions, such as anger or happiness. When exposed to events that trigger these emotions, they are likely to react with corresponding emotional responses. Strong Propensity for Dissemination: Upon encountering events that trigger specific emotions, Emotional Persons have a strong desire to spread the information to others. They aim to raise awareness and elicit empathy by sharing these events. Emotional Dissemination Behavior: In the process of sharing events, Emotional Persons tend to express their own emotions, which can enhance the spread of the events on social media platforms and potentially influence the emotions of other users. Role Transformation: Emotional Persons may shift between different events and corresponding emotions. For instance, they may express anger in response to an event like ``a teacher in Guangzhou physically punishing a student,'' whereas they may exhibit happiness in response to an event like ``a woman being detained for spreading rumors.'' In summary, Emotional Persons exert a certain level of influence on social media platforms. By sharing and expressing emotions, they drive the spread of specific events and emotions, and may also affect the emotions and viewpoints of other users. Understanding and paying attention to this group can help us better grasp the dynamics of public opinion on social media platforms.\\ \hline
			Gatekeeper & The ``Gatekeeper'' is a behavioral role that operates within the sphere of online social interactions. Individuals assuming this role exhibit a balanced rationality towards certain emotional types and are actively involved in the dissemination of events, while concurrently taking steps to diminish associated emotional responses. The primary characteristics of the Gatekeeper involve the rational management and propagation of emotions. They are adept at not only providing objective analyses of events but also at soothing negative sentiment within the digital space through the issuance of reassuring comments or the proactive sharing of information related to the event. For example, in the context of a societal event, a Gatekeeper might publish analyses of the event's objective facts on social media platforms such as Weibo or Twitter and simultaneously offer solace to individuals exhibiting anger, guiding them toward a more rational perspective. Thus, the Gatekeeper's function is to attenuate negative emotions throughout the dissemination process, contributing to the harmony and stability of the online environment. Overall, the behaviors of the Gatekeeper serve to modulate the strength of specific emotions transmitted during event propagation. By providing information that is more objective and calm, they may reduce the potent emotional reactions that an event could provoke, thereby shaping the collective emotional trajectory. Gatekeepers fulfill a critical role in the online ecosystem as emotional stabilizers, aiding in the preservation of online order and fostering a rational public understanding of events.\\ \hline
			Rationalist &The term ``Rationalist'' refers to an individual who exhibits a relatively rational response to certain emotional types in a specific event or situation. Rather than being readily triggered into emotional reactions by the event itself, they tend to approach the analysis and dissemination of information with an objective and rational mindset. This role is particularly prevalent on social media platforms, where individuals may participate in the spread of trending events without being swayed by emotions, maintaining a relatively calm and composed demeanor. For instance, in the dissemination of angry sentiment, Rationalists focus on the core aspects of the event, such as factual truths and relevant policies, and strive to remove emotional elements during the sharing process. They may help steer public discourse towards a focus on the event itself, rather than being influenced by emotions. Such a role contributes to the maintenance of rationality and fairness in public discussions and helps prevent adverse phenomena like cyberbullying. However, Rationalists do not always remain entirely rational. As an event unfolds, they may transition into other roles such as Emotional persons, Gatekeepers, or Onlookers. In the role of a Sensitizer, they might experience empathy or emotional fluctuations due to certain details of the event. As Gatekeepers, they could actively engage in the scrutiny and selection of information to ensure its authenticity and fairness. In the role of a Bystander, they may simply maintain an observant stance without offering opinions on the event. These shifts in roles reflect the psychological changes and coping strategies individuals employ in response to different situations. \\ \hline
			Onlooker & The ``Onlooker'' constitutes a demographic that adopts a passive observing position about online or social events. They are generally consumers of event-related information, paying attention to developments without contributing to the dissemination or discourse surrounding the event. This role is distinguished by a stance of neutrality, devoid of intense emotional bias towards the event, and a lack of initiative to participate in its discussion. Onlookers are inclined to follow the trajectory of events but seldom voice their opinions or engage in the redistribution of information. This segment of the population may shift to the role of Rationalist when confronted with particular events, suggesting that they may start to engage more intently and participate actively in discussions regarding the event under certain conditions, showing a more assertive demeanor. Overall, the Onlooker fulfills a role that maintains a certain distance and observes the dynamics of an event, yet they are not completely disengaged from social events. In specific contexts, they may transition into active participants or contributors to the event's discourse. This shift in roles illustrates the psychological and behavioral adaptations that individuals display when presented with varying event scenarios.\\ \hline
		\end{tabular}%
\end{table*}	

\section{Details of Corpora Construction and Datasets Statistics}
\label{Corpora}
\begin{figure}[htb]
	\centering
	\includegraphics[width=\linewidth]{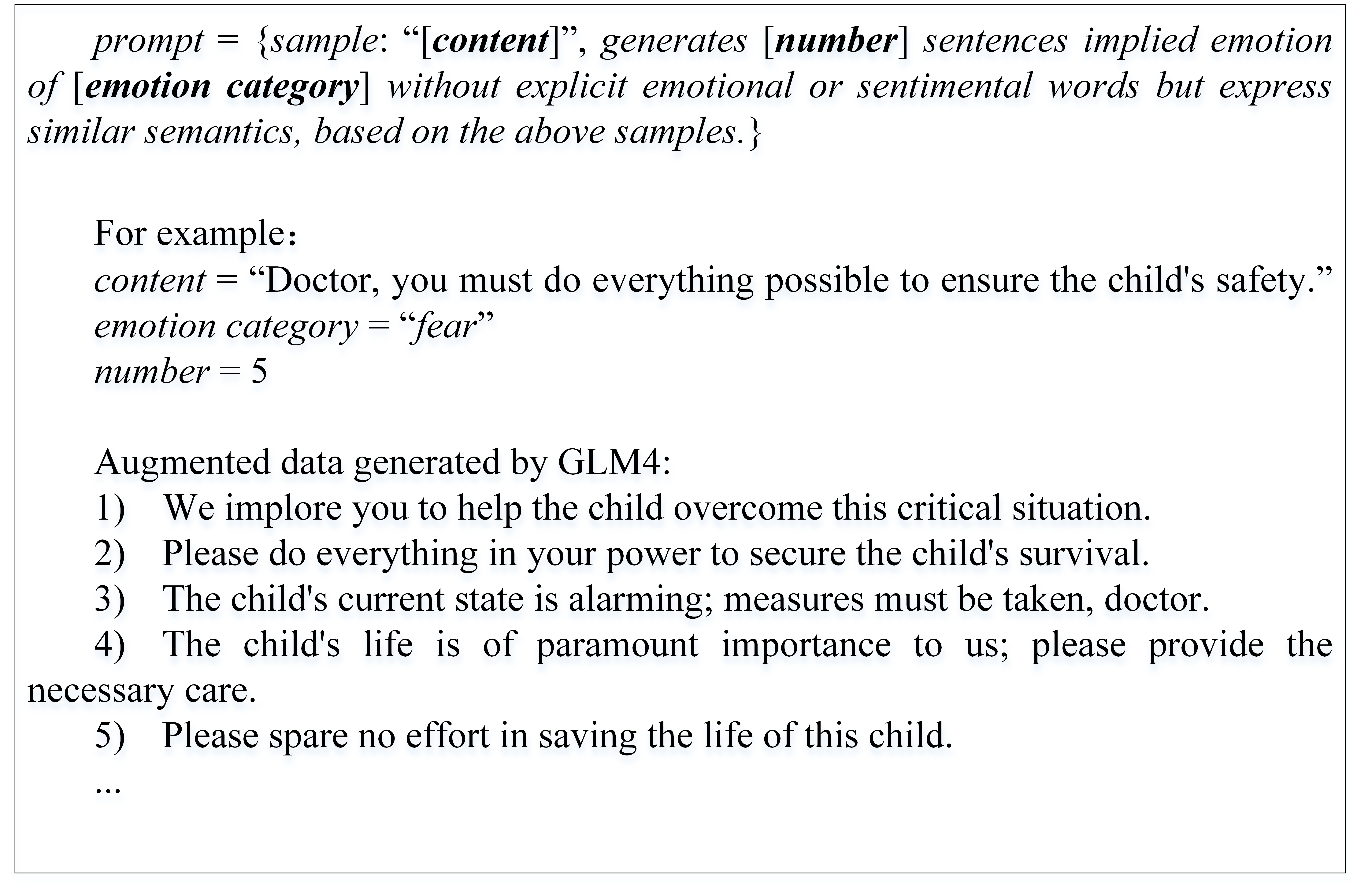} 
	\caption{The prompt for data augmentation.}
	\label{fig_dataaug}
\end{figure}

The original Chinese data in this study is collected from the popular social media  Weibo\footnote{\url{https://weibo.com/}} in China. We construct the English PIEA dataset based on the raw data from twitter100m\_tweets\footnote{\url{https://hf-mirror.com/datasets/enryu43/twitter100m_tweets}} and twitter100m\_users\footnote{\url{https://hf-mirror.com/datasets/enryu43/twitter100m_users}}. Compared to Weibo, the Twitter dataset does not include following relationships between users.
For the collected raw data, we first perform data anonymization and utilize two sentiment lexicons, Emotion Ontology\footnote{\url{https://github.com/ZaneMuir/DLUT-Emotionontology}} (in Chinese) and MPQA Subjectivity Lexicon\footnote{\url{https://mpqa.cs.pitt.edu/lexicons/subj_lexicon/}} (in English), to filter and establish implicit emotion datasets.
Four post-graduate student annotators independently label the data, achieving average inter-annotator agreement rates of 87.3\% and 79.8\% in Weibo and Twitter corpora, respectively. We randomly divide the datasets into training, validation, and testing sets in an 8:1:1 ratio. Due to the significant imbalance in the distribution of emotional data, we observe that existing models struggle with recognizing minor emotions like \textit{fear} and \textit{surprise}. Therefore, we construct class-balanced training datasets by sampling instances from each emotional category to create prompts and using the online LLM GLM4 to generate augmented data (the prompt for data augmentation is shown as Figure~\ref{fig_dataaug}), which is then integrated into the training set. Detailed statistics of the experimental datasets are presented in Table~\ref{Tab:dataset}.

\begin{table}[htb]
	\centering
	\small
	\setlength{\tabcolsep}{1.0mm}
	\caption{Detailed statistics of the experimental datasets, the number in () is the amount of augmented training data for each emotion category, \# means the number of corresponding items, R$_\text{following}$ represents the following relationships between users.}
	\label{Tab:dataset}
	\begin{tabular}{ccccccccc}
		\hline
		& \multicolumn{3}{c}{Weibo (Total:12384)}  & \multicolumn{3}{c}{Twitter (Total:16856)}  \\
		\cline{2-7}
		Emotion  & \begin{tabular}[c]{@{}c@{}}Train\\ (1500)\end{tabular} & Validate & Test & \begin{tabular}[c]{@{}c@{}}Train\\ (2000)\end{tabular} & Validate & Test  \\
		\hline
		Happy  & 1355 & 192 & 178 & 3920 & 490  & 490  \\
		Anger  & 451  & 54  & 65  & 986  & 123  & 117  \\
		Sad  & 919 & 107 & 104 & 1965 & 245  & 246  \\
		Disgust  & 360 & 35  & 45  & 1556  & 194  & 203 \\
		Fear & 23  & 10  & 10  & 530 & 66 & 65 \\
		Surprise & 198 & 30  & 23  & 780  & 98 & 92 \\
		Neutral  & 4188  & 514 & 517 & 1694 & 212 & 212 \\\hline
		Total & \begin{tabular}[c]{@{}c@{}}7494\\ (10500)\end{tabular} & 942 & 942 & \begin{tabular}[c]{@{}c@{}}11431\\ (14000)\end{tabular}  & 1428 & 1428\\ 
		\hline
		\# Users & & 3508 & & & 5676 &\\
		\# R$_\text{following}$& & 5927  & & &- &\\
		\hline
	\end{tabular}
\end{table}

\section{Details of the Baselines}
\label{Baselines}
To validate the advance of our model, we compare it with a range of SOTA sentiment/emotion analysis baselines, which can be divided into the following three categories.

\textbf{Methods based on knowledge enhancement}: \textbf{KGAN} \cite{zhong2023knowledge}, this technique integrates a knowledge graph to a multiview learning process for sentimental semantic representation.
\textbf{MMML} \cite{MMML24}, a multimodal multi-loss fusion network for sentiment analysis that optimizes the selection and fusion of multimodal knowledge encoders.
All these baselines are reproduced according to the optimal configuration given in the original paper.

\textbf{Methods based on GNNs}: \textbf{RDGCN} \cite{zhao2024rdgcn} is an enhanced dependency graph convolutional network that improves the calculation of importance in distance and type views of dependency relationships.
\textbf{DAGCN} \cite{wang2024dagcn}, a distance-based and aspect-oriented graph convolutional network which incorporates distance-based syntactic weights and sentimental aspect fusion attention.
All these baselines are also reproduced according to the original optimal configuration.

\textbf{Methods based on LLMs}: 
We employed two LLMs with different architectures as the foundation, both of which have demonstrated outstanding performance across various NLP tasks covering both Chinese and English. Building on this, we introduced two SOTA LLM-based semantic reasoning mechanisms to enhance the deep understanding capabilities of the LLM baselines.

\textbf{GLM4} \cite{glm2024chatglm} is an online LLM with tens of billions of parameters released by Zhipu, that combines a sophisticated transformer architecture with multimodal integration, enabling it to excel in long-context processing and multilingual support. We conduct the experiments by calling its API from \url{https://bigmodel.cn/}. 
\textbf{ChatGLM-6B}(ft) \cite{glm6b} is the open-source version of GLM series with 6B parameters, and has been fine-tuned using the training set. 
\textbf{Qwen-14B}(ft) \cite{qwen2.5}, Qwen2.5-14B is an open-source LLM released by Alibaba specifically optimized for logical reasoning and natural language understanding and has been fine-tuned using the training set. 
\textbf{THOR} \cite{fei2023} improves implicit sentiment analysis through a three-step chain-of-thought (COT) reasoning. \textbf{TOC} \cite{weinzierl-harabagiu-2024-tree} constructs a tree-of-counterfactual prompting, and enhance the reasoning ability of LLM through chains-of-explanation and chain-of-contrastive verification. 
We respectively integrate the THOR and TOC mechanisms with the LLMs mentioned above to enhance implicit emotion inference. See the Figures~\ref{fig_case1}-\ref{fig_case3} for detailed reasoning processes and prompts.
\textbf{DeepSeek-R1} \cite{deepseekr1} is the latest online LLM with 671 billion parameters trained via large-scale reinforcement learning without supervised fine-tuning as a preliminary step, demonstrating remarkable performance on reasoning. We conduct the experiments by calling its API provided by Aliyun\footnote{\url{https://www.aliyun.com/}}.

\section{Detailed Reasoning Processes for Case Study}
\label{detailcase}
\begin{figure*}[htb]
	\centering
	\includegraphics[width=0.9\textwidth]{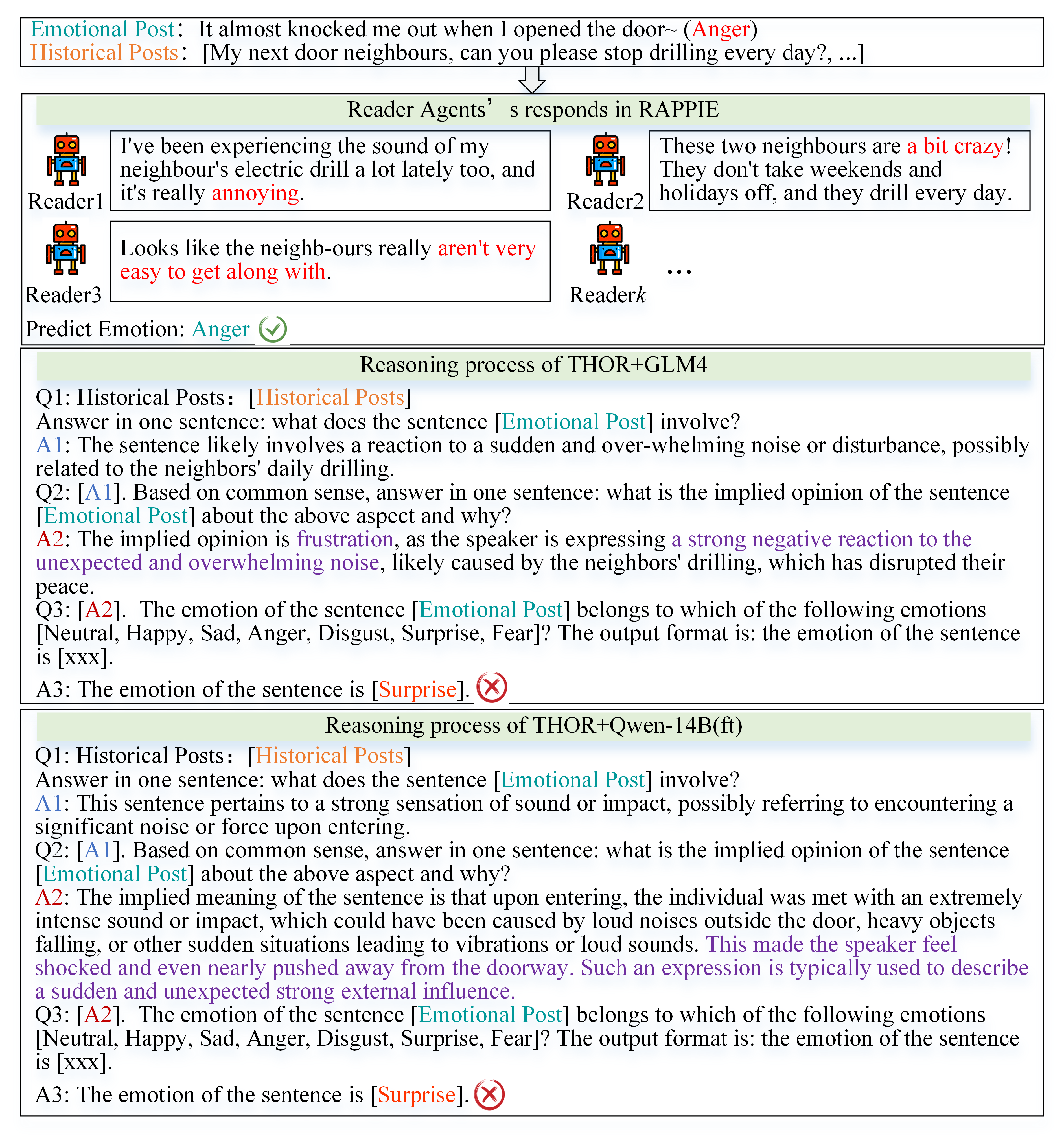} 
	\caption{A case study of the prediction process for RAPPIE and different LLMs with THOR.}
	\label{fig_case1}
\end{figure*}

\begin{figure*}[htb]
	\centering
	\includegraphics[width=0.9\textwidth]{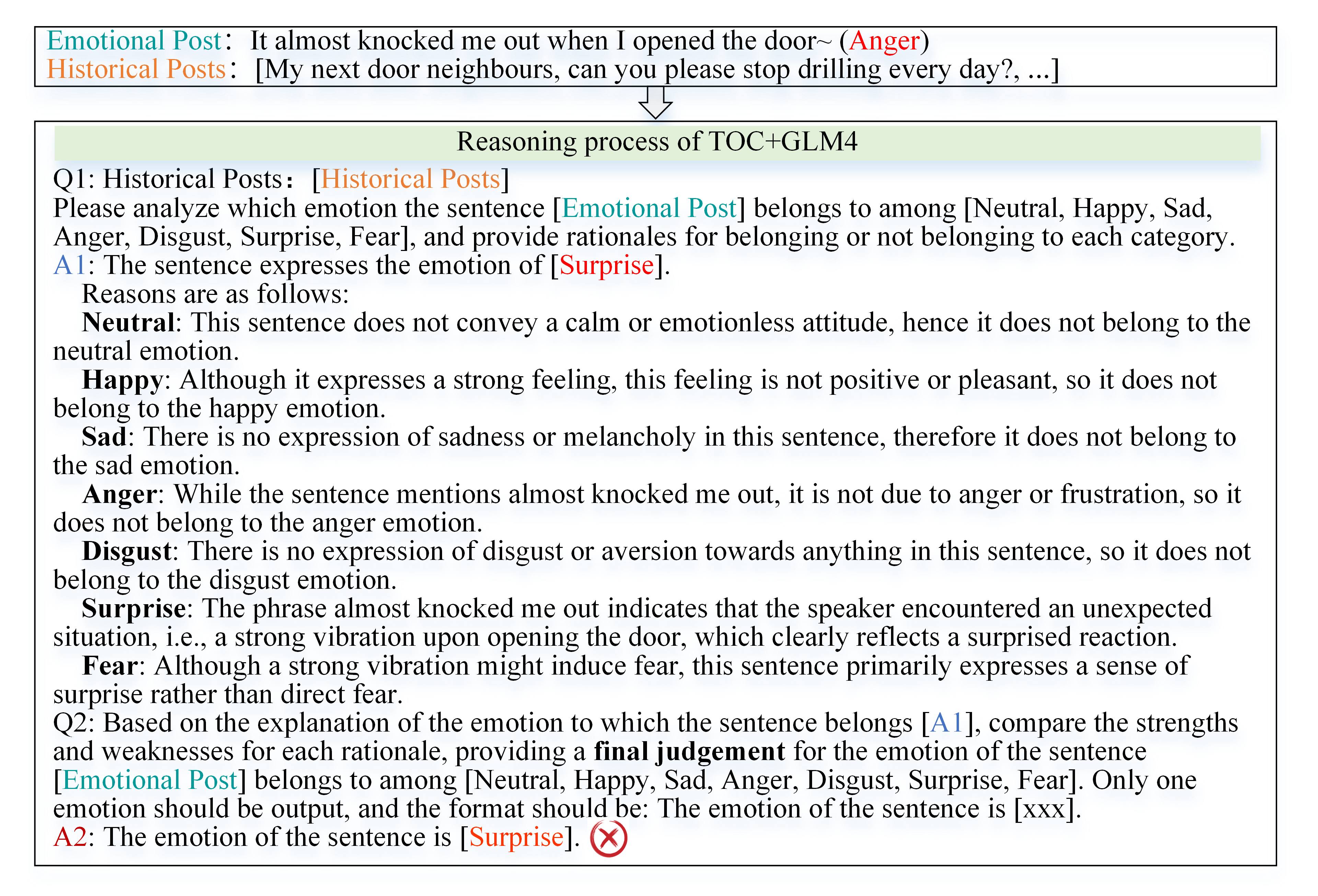} 
	\caption{A case study of the prediction process for TOC+GLM4.}
	\label{fig_case2}
\end{figure*}

\begin{figure*}[htb]
	\centering
	\includegraphics[width=0.9\textwidth]{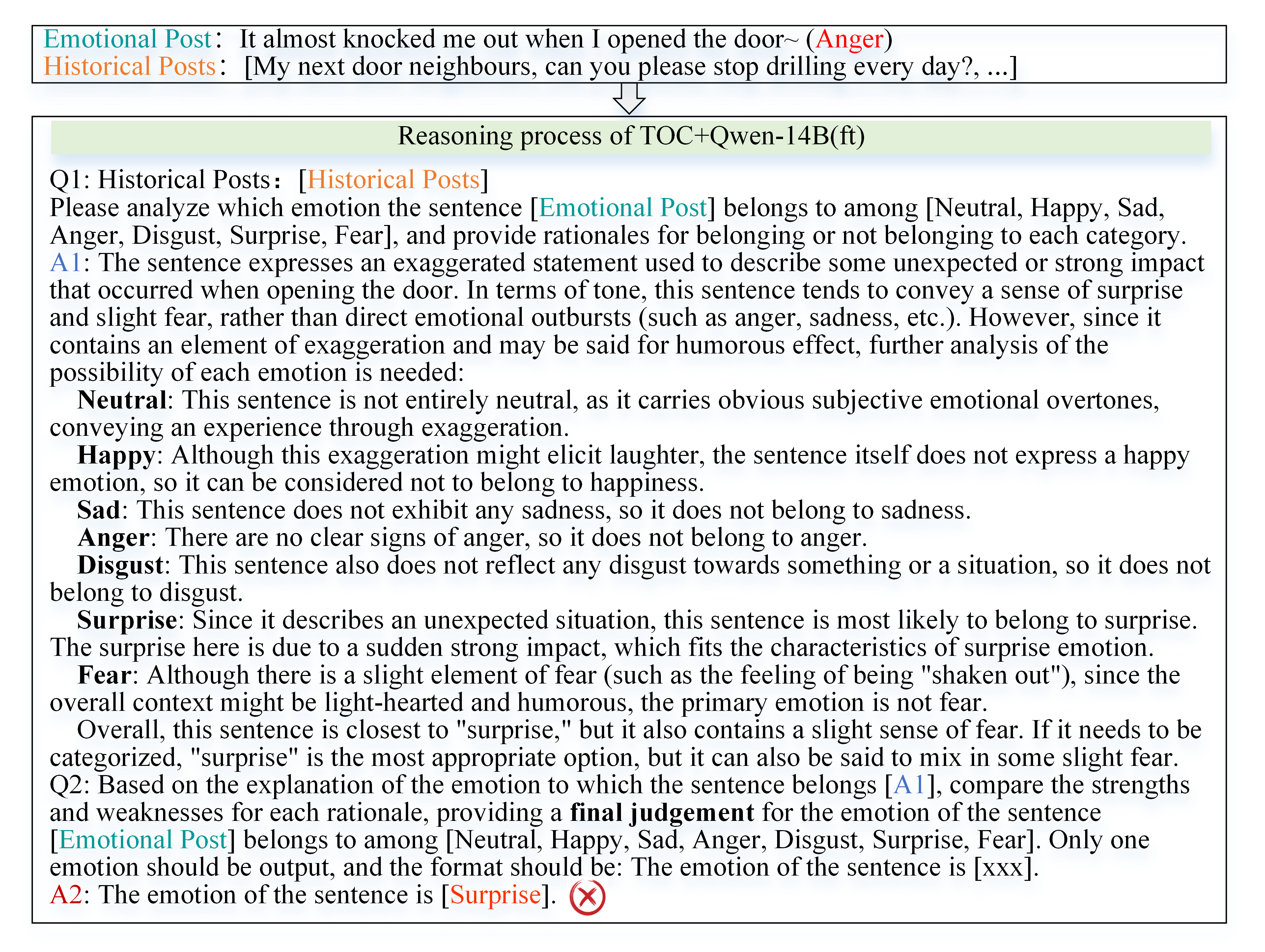} 
	\caption{A case study of the prediction process for TOC+Qwen-14B(ft).}
	\label{fig_case3}
\end{figure*}

We provide an instance that illustrates detailed reasoning processes for a given content and its associated historical posts list, shown in Figures~\ref{fig_case1}-\ref{fig_case3}.

\end{document}